\documentclass[11pt]{article}

\usepackage[preprint]{acl}
\makeatletter
\AtBeginDocument{\@ifundefined{@LN@makecol}{}{\let\@makecol\@LN@makecol}}
\makeatother

\usepackage{times}
\usepackage{latexsym}

\usepackage[T1]{fontenc}

\usepackage[utf8]{inputenc}

\usepackage{microtype}

\usepackage{inconsolata}

\usepackage{graphicx}
\graphicspath{{figs/}}

\usepackage{amsmath,amssymb}
\usepackage{booktabs}
\usepackage{colortbl}
\definecolor{oursbg}{RGB}{232,240,252}
\definecolor{bestbg}{RGB}{210,226,250}
\usepackage{multirow}
\usepackage{xcolor}
\usepackage{tikz}\usetikzlibrary{calc,positioning}
\usepackage{algorithm,algpseudocode}
\usepackage{listings}
\usepackage{float}
\usepackage{placeins}   %
\usepackage{enumitem}
\newcommand{\thinktag}{\texttt{<think>}}
\newcommand{\closetag}{\texttt{</think>}}
\newcommand{\zb}{z_{\mathrm{base}}}
\newcommand{\zc}{z_{\mathrm{brief}}}
\newcommand{\zv}{z_{\mathrm{verbose}}}
\newcommand{\ci}[2]{{\scriptsize$[#1,#2]$}}

\newcommand{\method}{DoubtLess}   %
\newcommand{\promptbox}[2]{%
  \par\medskip\noindent\begin{tikzpicture}
  \node[draw=black!35, fill=black!4, rounded corners=3pt, inner xsep=7pt, inner ysep=6pt, text width=\dimexpr\linewidth-14pt-0.8pt\relax, align=flush left]
  {{\small\sffamily\bfseries #1}\par\smallskip{\small #2}};
  \end{tikzpicture}\par\medskip}

\newcommand{\chk}[1]{\textcolor{red!75!black}{#1}}
\newcommand{\om}{{[}.\kern0.1em.\kern0.1em.{]}}  %
\newcommand{\dneg}[1]{{\setlength{\fboxsep}{0.5pt}\colorbox{red!25}{\vphantom{Ag}#1}}}
\newcommand{\dpos}[1]{{\setlength{\fboxsep}{0.5pt}\colorbox{green!30}{\vphantom{Ag}#1}}}
\newcommand{\casebox}[4]{%
  \noindent\begin{tikzpicture}
  \node[draw=black, line width=0.5pt, fill=#3, inner xsep=5pt, inner ysep=4pt, text width=\dimexpr\linewidth-10pt-1pt\relax, align=flush left]
  {{\small\bfseries #1}\par
   {\footnotesize\itshape #2}\par\vspace{2pt}
   {\footnotesize\setlength{\parskip}{1.2pt}#4\par}};
  \end{tikzpicture}\par}
\newcommand{\exPlainTok}{3,414}
\newcommand{\exOursTok}{1,461}

\title{\method{}: Training-Free Contrastive Decoding for Efficient Reasoning}

\author{Minghao Luo \\
  The Chinese University of Hong Kong \\
  \texttt{luominghao@link.cuhk.edu.hk} \\\And
  Liang Chen\thanks{Corresponding author.} \\
  EPFL \\
  \texttt{l.chen@epfl.ch} \\}

\begin{document}
\maketitle

\begin{abstract}
Reasoning models achieve strong performance through long chains of thought (CoT), but often spend more tokens than needed to solve a problem.
Existing ways to shorten the CoT require additional training for each model, add a fixed steering vector to the model's hidden states, or use decoding heuristics such as stopping early or banning words like \emph{wait}, which can reduce accuracy.
We propose \method{}, a training-free decoding method in which a much smaller reasoning model steers the large one towards brevity.
The small model reads the large model's CoT so far twice, once prompted to be brief (the \emph{brief thinker}) and once prompted to be verbose (the \emph{verbose thinker}), and the difference between its two logit vectors, scaled by a strength, is added to the large model's logits at every step of the CoT.
On seven benchmarks of competition mathematics, graduate-level science and competitive programming, \method{} shortens the CoT of Qwen3-32B by 31\% with unchanged accuracy, outperforming five inference-time baselines.
We also show that \method{} reduces thinking tokens per turn by 20\% in a self-debugging coding agent with unchanged task performance.
It also generalizes to two other model families with unchanged accuracy.
When combined with training-based compression, it further shortens the CoT of a model already trained for brevity.
Ablations and behavioral analysis show that \method{} reduces repeated checking, with most token savings coming after the correct answer first appears in the CoT.
Our code is available at \url{https://github.com/leoluolol/DoubtLess}.
\end{abstract}

\section{Introduction}\label{sec:intro}
\begin{figure}[t]
\centering
\includegraphics[width=\columnwidth]{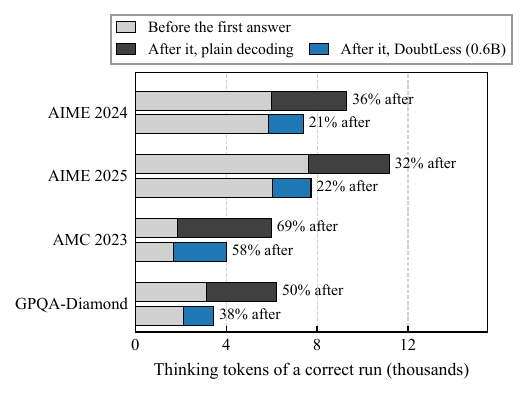}
\caption{Qwen3-32B: mean thinking tokens of a correct run before and after the correct answer first appears in the CoT, under plain decoding (the model as released) and under \method{}. Of the tokens \method{} removes, 65--89\% come after the answer (Table~\ref{tab:whereall}).}
\label{fig:motivation}
\end{figure}
Reasoning models such as Qwen3 and DeepSeek-R1 write a chain of thought (CoT) between the tokens \thinktag{} and \closetag{} before they answer.
On a hard problem the CoT holds most of the output, and a large part of it re-derives and re-verifies a result the model has already reached~\citep{chen2024overthinking,sui2025stopoverthinking}.
Under plain decoding, that is, sampling from the model as released, a correct run of Qwen3-32B writes 32--69\% of its CoT after the correct answer first appears (Figure~\ref{fig:motivation}).
Not all of this checking is redundant. A check made while the model works, or a first check of its answer, can catch an error, but repeating a check that has already been made adds tokens without adding information.
A method that shortens the CoT should remove the redundant checks and keep the necessary ones.

Existing methods for shortening the CoT fall into two groups.
Training-based methods add a length penalty or a token limit to reinforcement learning~\citep{aggarwal2025l1,hou2025thinkprune,luo2025o1pruner} or fine-tune on short traces~\citep{chen2024overthinking}. Both change the weights, so every new model needs another round of training.
Inference-time methods ask for brevity in the prompt~\citep{renze2024ccot,han2024tale,xu2025chainofdraft}, block words such as \emph{wait} that open another round of checking~\citep{wang2025nowait}, end the CoT once an answer looks ready or a token budget is spent~\citep{yang2025deer,song2025thinkbrake,muennighoff2025s1}, or add a fixed steering vector to the model's hidden states~\citep{azizi2026asc,chen2025seal,lin2025thinkingspeed}.
Prompting depends on how closely the model follows the instruction.
Keyword bans suppress the same words across contexts, activation steering adds the same vector to the hidden states at every step, and early stopping ends the CoT once its criterion is met.
These controls can shorten useful reasoning as well as redundant checks.

\begin{figure}[t]
\centering
\includegraphics[width=\columnwidth]{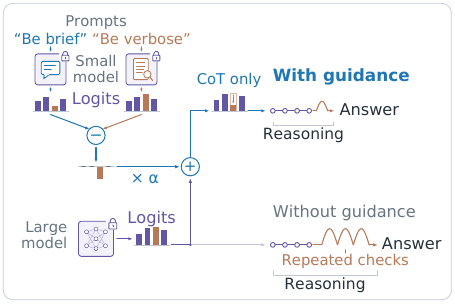}
\caption{\method{}. A small model that shares the tokenizer reads the large model's CoT so far twice, once prompted to be brief (brief thinker) and once prompted to be verbose (verbose thinker); the difference between the two logit vectors, scaled by $\alpha$, is added to the large model's logits while it writes the CoT.}
\label{fig:schematic}
\end{figure}

In this paper we propose \method{}, a training-free decoding method that steers a reasoning model with a much smaller reasoning model that shares its tokenizer (Figure~\ref{fig:schematic}).
At every step of the CoT, the small model reads the text the large model has written so far twice: once after an instruction to go straight to the answer, as the \emph{brief thinker}, and once after an instruction to double-check every step, as the \emph{verbose thinker}.
The difference between the two logit vectors, multiplied by a strength $\alpha$, is added to the large model's logits.
The small model only suggests a direction, towards what the brief thinker would write next and away from what the verbose thinker would.
The large model still decides which tokens are possible, and the CoT can end only where the large model itself would consider ending it (Section~\ref{sec:method}).
Inspired by DExperts~\citep{liu2021dexperts}, \method{} steers the large model through the vocabulary it shares with a small model, but needs no trained expert, since both sides of the contrast come from one off-the-shelf small model under two instructions.

Two properties set \method{} apart from prompting, keyword bans, activation steering and early stopping.
First, the contrast is dynamic. It is recomputed from the current context at every step, so a word such as \emph{But} is pushed down after an answer and left almost untouched before it, whereas a keyword ban blocks it everywhere and activation steering applies one fixed vector at every step.
Second, the brief and verbose thinkers disagree most after the correct answer first appears in the CoT, where the brief thinker would wrap up and the verbose one would check again (Section~\ref{sec:where}).
Most of the removed tokens come from this later checking.

We compare \method{} with five inference-time baselines on seven benchmarks of competition mathematics, graduate-level science and competitive programming, with the model's recommended sampling settings and three seeds per problem.
With Qwen3-32B steered by Qwen3-0.6B, \method{} removes 31\% of the thinking tokens with unchanged accuracy, more than any baseline.
It keeps accuracy where the baselines lose it. On AIME 2025, for example, blocking the words that open another round of checking costs 14 points, while \method{} costs none.
\method{} also extends to an agent, a model trained for brevity and two other model families.
It reduces thinking per turn by 20\% in a self-debugging coding agent and further shortens the CoT of a model trained for brevity with LACONIC~\citep{liu2026laconic}.
On AIME 2025 and AMC 2023, it removes 25\% and 40\% of the thinking of EXAONE-4.0 and Granite-4.2, each steered by a small model of its own family, where a concise prompt removes 8\% and 14\%.
Finally, our ablations show that the strength sets how much is removed, that a negative strength lengthens the CoT instead, and that the shortening depends on which tokens the small model pushes down.

Our main contributions are as follows:
\begin{itemize}[leftmargin=1.1em,nosep]
\item \method{}, a training-free decoding method in which an off-the-shelf small model, run under a brief and a verbose instruction, supplies a logit direction that shortens a reasoning model's CoT.
\item An evaluation in which \method{} removes more thinking than five inference-time baselines on seven benchmarks with unchanged accuracy, and shortens the CoT of a self-debugging coding agent and of two other model families.
\item An analysis showing that the two thinkers disagree mainly after the answer first appears in the CoT, which is where most of the removed tokens lie.
\end{itemize}

\section{Method}\label{sec:method}

\subsection{Preliminaries}\label{sec:prelim}
\paragraph{Decoding a reasoning model.}
Given a prompt $x$, a language model $M$ with vocabulary $\mathcal{V}$ generates a response $y=(y_1,\dots,y_n)$ one token at a time,
\begin{equation}
  p_M(y\mid x)\;=\;\prod_{t=1}^{n} p_M(y_t\mid x,y_{<t}).
  \label{eq:ar}
\end{equation}
At each step it returns a logit vector $z_t\in\mathbb{R}^{|\mathcal{V}|}$, and its next-token distribution at temperature $T$ is
\begin{equation}
  p_M^{T}(\cdot\mid x,y_{<t})\;=\;\mathrm{softmax}(z_t/T),
  \label{eq:temp}
\end{equation}
with $p_M=p_M^{1}$.
A reasoning model writes its CoT between \thinktag{} and \closetag{}, then its answer.

\paragraph{Sampling.}
Sampling draws $y_t$ from Equation~\eqref{eq:temp}, usually truncated to the most probable tokens.
The top-$p$ set $\mathcal{V}_p(q)$ of a distribution $q$ over $\mathcal{V}$ is the smallest set of tokens such that~\citep{holtzman2020nucleus}
\begin{equation}
  \textstyle\sum_{v\in\mathcal{V}_p(q)} q(v)\;\geq\; p,
  \label{eq:topp}
\end{equation}
and the top-$k$ set holds the $k$ most probable tokens.
Truncating to a set $S$ sets every other logit to $-\infty$~\citep{liu2021dexperts},
\begin{equation}
  [z]_S(v)\;=\;\begin{cases} z(v) & \text{if } v\in S,\\ -\infty & \text{otherwise,}\end{cases}
  \label{eq:mask}
\end{equation}
so that the softmax renormalizes over $S$.
A sampler applies temperature, top-$k$ and top-$p$ in turn. For thinking, Qwen3 recommends $0.6$, $20$ and $0.95$.

\subsection{\method{}}\label{sec:contrast}
Let $D$ be a much smaller reasoning model that shares the tokenizer of $M$ (Figure~\ref{fig:schematic}), and write $\zb$ for the logits of $M$ at the current step.

\begin{figure}[t]
\input{examples/direction}
\caption{Before the answer, the direction leaves the CoT nearly untouched. After it, the words that open another check are pushed down. The CoT that plain decoding writes for AIME 2025 problem 1, read by Qwen3-0.6B under both instructions. \textbf{\textcolor{red!75!black}{Red}}: down-weighted; \textbf{\textcolor{green!45!black}{green}}: up-weighted (threshold in Appendix~\ref{app:where}); \om{} marks omitted text.}
\label{fig:direction}
\end{figure}
\begin{table*}[t]
\centering\footnotesize
\caption{\method{} removes the most thinking tokens overall. Qwen3-32B, three seeds per problem. Plain decoding: accuracy (\%) and mean thinking tokens; other rows: change against plain decoding, in points and in \% (intervals in Table~\ref{tab:full}). Overall: the 528 problems of the seven benchmarks together. Bold and underlined: largest and second-largest removal among the six methods. Agent: pass rate and thinking tokens per turn on HumanEval+.}
\label{tab:main}
\setlength{\tabcolsep}{2pt}\renewcommand{\arraystretch}{1.05}
\begin{tabular}{ll rrrrr r r r r}
\toprule
 & & \multicolumn{5}{c}{Mathematics} & \multicolumn{1}{c}{Science} & \multicolumn{1}{c}{Code} & \multicolumn{1}{c}{Agent} & \multicolumn{1}{c}{Overall} \\
\cmidrule(lr){3-7}\cmidrule(lr){8-8}\cmidrule(lr){9-9}\cmidrule(lr){10-10}\cmidrule(lr){11-11}
 & & \multicolumn{1}{c}{\shortstack{AIME\\2024}} & \multicolumn{1}{c}{\shortstack{AIME\\2025}} & \multicolumn{1}{c}{\shortstack{HMMT\\2025}} & \multicolumn{1}{c}{\shortstack{AMC\\2023}} & \multicolumn{1}{c}{\shortstack{Olympiad-\\Bench}} & \multicolumn{1}{c}{\shortstack{GPQA-\\Diamond}} & \multicolumn{1}{c}{\shortstack{LiveCode-\\Bench}} & \multicolumn{1}{c}{\shortstack{HumanEval+\\agent}} & \multicolumn{1}{c}{\shortstack{All\\seven}} \\
\midrule
 & Acc. & 81.1 & 72.2 & 53.3 & 96.7 & 70.7 & 68.7 & 66.3 & 88.4 & 70.8 \\
\multirow{-2}{*}{Plain decoding} & Tokens & 12,152 & 15,606 & 16,817 & 6,130 & 12,246 & 7,354 & 11,511 & 2,718 & 10,254 \\
\midrule
 & Acc.\ change & $+2.2$ & $+3.3$ & $-6.7$ & $-0.8$ & $-2.0$ & $-1.0$ & $-0.7$ & $+1.4$ & $-1.0$ \\
\multirow{-2}{*}{Concise prompt} & Token change & {\bfseries\boldmath $-30\%$} & \underline{$-28\%$} & $-13\%$ & \underline{$-31\%$} & \underline{$-25\%$} & \underline{$-40\%$} & $-8\%$ & $-13\%$ & \underline{$-25\%$} \\
\addlinespace[2pt]
 & Acc.\ change & $-3.3$ & $-14.4$ & $-6.7$ & $-2.5$ & $-1.0$ & $-0.3$ & $-4.0$ & $-0.8$ & $-2.7$ \\
\multirow{-2}{*}{NoWait} & Token change & $-21\%$ & $-23\%$ & \underline{$-19\%$} & $-24\%$ & $-17\%$ & $-21\%$ & \underline{$-18\%$} & \underline{$-29\%$} & $-20\%$ \\
\addlinespace[2pt]
 & Acc.\ change & $+1.1$ & $+2.2$ & $0.0$ & $+0.8$ & $+0.7$ & $0.0$ & $-4.7$ & $+4.5$ & $-0.5$ \\
\multirow{-2}{*}{DEER} & Token change & $-8\%$ & $-10\%$ & $-6\%$ & $-20\%$ & $-12\%$ & $-2\%$ & $-17\%$ & {\bfseries\boldmath $-41\%$} & $-10\%$ \\
\addlinespace[2pt]
 & Acc.\ change & $-1.1$ & $0.0$ & $-1.1$ & $+0.8$ & $-3.3$ & $-2.0$ & $-4.7$ & $-6.9$ & $-2.3$ \\
\multirow{-2}{*}{ThinkBrake} & Token change & $-4\%$ & $-3\%$ & $-4\%$ & $-3\%$ & $-4\%$ & $-37\%$ & {\bfseries\boldmath $-44\%$} & $-27\%$ & $-21\%$ \\
\addlinespace[2pt]
 & Acc.\ change & $+1.1$ & $+1.1$ & $-2.2$ & $+1.7$ & $-1.7$ & $+1.2$ & $-0.3$ & $-1.4$ & $+0.2$ \\
\multirow{-2}{*}{ASC} & Token change & $-12\%$ & $-8\%$ & $+3\%$ & $-11\%$ & $-4\%$ & $-4\%$ & $-4\%$ & $-5\%$ & $-5\%$ \\
\addlinespace[2pt]
\rowcolor{oursbg} & Acc.\ change & $+2.2$ & $0.0$ & $+3.3$ & $+0.8$ & $-1.7$ & $-2.4$ & $-0.3$ & $+2.0$ & $-0.9$ \\
\rowcolor{oursbg}\multirow{-2}{*}{\shortstack[l]{\textbf{\method{}}\\\textbf{(0.6B)}}} & Token change & \underline{$-27\%$} & {\bfseries\boldmath $-29\%$} & {\bfseries\boldmath $-31\%$} & {\bfseries\boldmath $-34\%$} & {\bfseries\boldmath $-27\%$} & {\bfseries\boldmath $-46\%$} & $-16\%$ & $-20\%$ & {\bfseries\boldmath $-31\%$} \\
\bottomrule
\end{tabular}
\end{table*}

\paragraph{Definition.}
At every decoding step, $D$ reads the same prefix as $M$ (the problem, the instruction to reason step by step, and the tokens generated so far) under two versions of the instruction that differ in one trailing sentence.
As the \emph{brief thinker} it is prompted to go straight to the answer without re-deriving or re-verifying anything. As the \emph{verbose thinker} it is prompted to double-check every step and to consider alternatives (the two sentences are given in Appendix~\ref{app:prompts}).
With $\zc$ and $\zv$ for the two resulting logit vectors, the next token is sampled from
\begin{equation}
  z \;=\; \zb \;+\; \alpha\,(\zc-\zv),
  \label{eq:gap}
\end{equation}
restricted to the top-$p$ set of $\mathrm{softmax}(\zb)$, the large model's distribution at temperature one (Equations~\eqref{eq:topp} and~\eqref{eq:mask}).
We call $\zc-\zv$ the \emph{direction} of the small model: the way its next-token expectations move when its instruction changes from thorough to brief.

\paragraph{What the direction does.}
Sampling from Equation~\eqref{eq:gap} at temperature $T$ draws from the large model's own distribution at that temperature, $p_M^{T}$ (Equation~\eqref{eq:temp}), reweighted by the ratio of the brief thinker's probability for the next token to the verbose thinker's:
\begin{equation}
  p(v)\;\propto\; p_M^{T}(v\mid x,y_{<t})\,
  \Bigl[\tfrac{p_D(v\mid x,y_{<t},\,\text{brief})}{p_D(v\mid x,y_{<t},\,\text{verbose})}\Bigr]^{\alpha/T}
  \label{eq:poe}
\end{equation}
(a difference of logits is a difference of log-probabilities up to a constant shared by all tokens, and the temperature divides the whole of $z$; Appendix~\ref{app:ratio}).
A token that the brief thinker expects more than the verbose thinker becomes more likely, one that the verbose thinker expects more becomes less likely, and where the two thinkers agree nothing changes.

\paragraph{When it acts.}
The two thinkers read the same text and differ only in the instruction, so they disagree only where the instruction matters (one CoT in Figure~\ref{fig:direction}; the analysis in Section~\ref{sec:where}).
While the model is still working towards an answer, both expect the derivation to continue, and the CoT is left almost unchanged: in the upper box of Figure~\ref{fig:direction} only two tokens are shaded, and the \emph{Wait} that opens a check inside the derivation is not one of them.
Once an answer has been stated, the verbose thinker expects another round of checking and the brief thinker expects the CoT to end, so the words that start a new check, such as \emph{But}, \emph{Let} and \emph{maybe} in the lower box, are down-weighted by a factor of five or more.
Unlike NoWait, \method{} therefore bans no word. It makes another round of checking less likely to begin once an answer is in place, and what the CoT loses is those rounds.
Whether the next sentence starts another check or continues the derivation is largely a matter of form, which a small model can judge even on a problem it could not solve.

\paragraph{Where the contrast applies.}
The contrast acts only on the CoT. The answer after \closetag{} is decoded by the large model alone.
Within the CoT, the large model keeps its own logits for \closetag{} and the end-of-sequence token and its own top-$p$ candidates, so the small model can only reorder tokens the large model was already willing to write (Algorithm~\ref{alg:step}, Appendices~\ref{app:setup} and~\ref{app:ratio}).

\section{Experimental setup}\label{sec:setup}
\paragraph{Models.}
The large model $M$ is Qwen3-32B~\citep{yang2025qwen3} in bfloat16 and the small model $D$ is Qwen3-0.6B. We write \method{} (0.6B) for this pair.
Section~\ref{sec:families} repeats the comparison with plain decoding and the concise prompt on two other families: EXAONE-4.0-32B steered by EXAONE-4.0-1.2B~\citep{lgai2025exaone4} and Granite-4.2-30B steered by Granite-4.2-3B~\citep{ibm2026granite42}.
Section~\ref{sec:ablation} also uses Qwen3-1.7B and Qwen3-4B as the small model.

\paragraph{Benchmarks.}
We use seven benchmarks that are the standard testing ground of reasoning models: four mathematics competitions, AIME 2024 and AIME 2025 (30 problems each;~\citealp{maa_aime}), HMMT February 2025 (30;~\citealp{hmmt2025}) and AMC 2023 (40;~\citealp{maa_amc}); graduate-level science questions from GPQA-Diamond (198;~\citealp{rein2024gpqa}); olympiad problems from OlympiadBench (100, drawn round-robin across subjects;~\citealp{he2024olympiadbench}); and programming problems from LiveCodeBench (a fixed subset of 100 from release 6, drawn round-robin across the three difficulty levels;~\citealp{jain2025livecodebench}).
Every benchmark is scored with its published answer key, and LiveCodeBench by running the program against the test cases (Appendix~\ref{app:exp}).

\paragraph{Baselines.}
We compare \method{} with plain decoding and with five training-free methods of the kinds in Section~\ref{sec:intro}: a prompt, a ban on words, two rules that end the CoT early and a steering vector.
\textbf{Concise prompt} asks the large model itself to be brief: the brief thinker's sentence, which tells the model to go straight to the answer and not to re-derive or re-verify anything, is appended to the large model's own instruction. How much it removes depends on how closely the large model follows the instruction.
\textbf{NoWait}~\citep{wang2025nowait} blocks 17 words that open another round of checking (\emph{wait}, \emph{alternatively}, \emph{hmm}, \emph{but}, \emph{however} and others) throughout the generation, by adding a bias of $-100$ to the logits of their single-token forms (89 token ids).
\textbf{DEER}~\citep{yang2025deer} ends the CoT early. At each of the first ten occurrences of \emph{Wait} it induces a trial answer, ends the CoT if the geometric mean of the trial's token probabilities exceeds $0.95$, and otherwise discards the trial and lets the CoT continue. It also ends the CoT once it reaches 0.8 of the budget.
A DEER run is a prefix of the plain-decoding run followed by \closetag{} and a freshly sampled answer, so we evaluate it on the stored plain-decoding runs with the same engine, sampler and seeds (Appendix~\ref{app:impl}).
\textbf{ThinkBrake}~\citep{song2025thinkbrake} ends the CoT at a sentence boundary once \closetag{} comes within a logit margin of $0.1$ of the most likely token, read after a single newline as in the authors' code, and is evaluated like DEER (Appendix~\ref{app:impl}).
\textbf{ASC}~\citep{azizi2026asc} adds a fixed steering vector to the large model's hidden states after one layer, at every decoding step. We use the training-free version in the authors' released code, in which the vector is the mean difference between the hidden states at the last token of a verbose and of a concise CoT, here Qwen3-32B's own CoTs on 100 MATH-500 problems~\citep{hendrycks2021math,lightman2024verify} under its plain instruction and under the brief thinker's sentence. It is added after layer 57 of 64, the layer the authors use for the 64-layer QwQ-32B, with the strength chosen on AIME 2026~\citep{maa_aime26} as for \method{} (Appendix~\ref{app:impl}).
We also decode the random-target control of Section~\ref{sec:control}.

\paragraph{Implementation details.}
All methods sample with Qwen3's recommended thinking settings, temperature $0.6$, top-$p$ $0.95$ and top-$k$ $20$, under a thinking budget of 32{,}768 tokens (24{,}576 on GPQA-Diamond). A run that reaches the budget without ending its CoT counts as wrong.
Every problem is decoded with three seeds. We call one decoding of one problem under one seed a \emph{run}, and accuracy is the mean over runs.
\method{} uses $\alpha{=}5$ (the best trade-off on AIME 2026; Appendix~\ref{app:alpha26}), one value for every model pair and every benchmark. Section~\ref{sec:strength} varies it.
All methods run in the same vLLM engine, with \method{} and the control computed inside it (Appendix~\ref{app:impl}).

\paragraph{Evaluation.}
We report accuracy and the change in the number of thinking tokens, the tokens of the CoT, against plain decoding.
The change is taken over all problems, solved or not, with a 95\% bootstrap interval over problems. Appendix~\ref{app:full} gives the formula and the change over the problems both methods solve, the measure of earlier work.
Since a run that reaches the budget counts as wrong, we also report the share of such runs.

\section{Results}\label{sec:results}
\begin{figure}[t]
\centering
\includegraphics[width=\columnwidth]{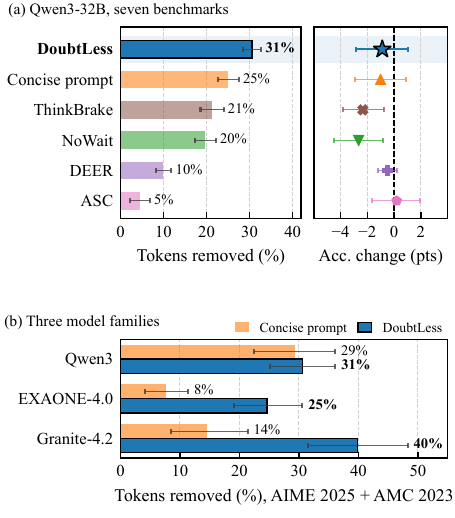}
\caption{\method{} removes the most thinking tokens with unchanged accuracy. (a) Percentage of thinking tokens removed and accuracy change against plain decoding, for Qwen3-32B on the seven benchmarks. (b) When evaluated on other model families, \method{} removes far more than the concise prompt (AIME 2025 and AMC 2023, each large model steered by a small model of its own family). Three seeds per problem, 95\% intervals (Tables~\ref{tab:full} and~\ref{tab:families}).}
\label{fig:tradeoff}
\end{figure}

\begin{figure*}[t]
\centering
\includegraphics[width=\textwidth]{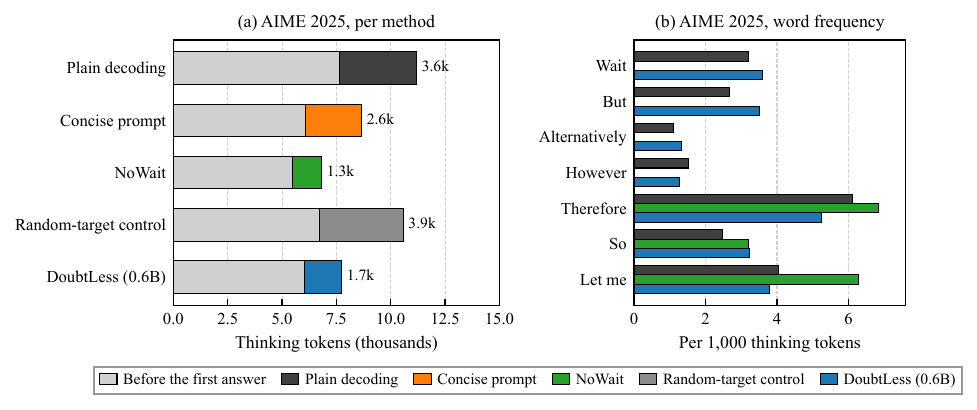}
\caption{\method{} removes mostly the thinking after the first answer, and the words of hesitation stay. AIME 2025. (a) Mean thinking tokens of a correct run before (light) and after (dark) the first occurrence of the correct answer, per method; number: tokens after it. (b) Selected words per thousand thinking tokens.}
\label{fig:removed}
\end{figure*}

\subsection{Main results}\label{sec:main}
\paragraph{\method{} shortens the CoT more than any baseline.}
Across the seven benchmarks, \method{} removes 31\% of the thinking tokens with unchanged accuracy (Table~\ref{tab:main}, overall column; intervals in Table~\ref{tab:full}).
This is the largest reduction among the six methods (Figure~\ref{fig:tradeoff}a).
The concise prompt comes closest, but \method{} removes more than it on six of the seven benchmarks.
The other baselines remove less, and ThinkBrake and NoWait lose more accuracy than \method{}.

\paragraph{The reduction holds on harder problems.}
\method{} shortens the CoT significantly on every benchmark (Table~\ref{tab:full}).
The baselines are uneven.
On mathematics, the concise prompt removes less on harder benchmarks, from 31\% on AMC 2023 to 13\% on HMMT 2025, the easiest and the hardest for plain decoding, while \method{} still removes 31\% on HMMT 2025.
ThinkBrake cuts the CoT on science and code but barely on mathematics.

\paragraph{\method{} keeps accuracy on every benchmark.}
In contrast, NoWait, DEER and ThinkBrake each lose accuracy significantly on at least one benchmark (Table~\ref{tab:full}): NoWait on AIME 2025 and LiveCodeBench, DEER on LiveCodeBench, and ThinkBrake on OlympiadBench and LiveCodeBench.
\method{} also lowers the share of runs that reach the budget, from 2.4\% under plain decoding to 0.4\%.

\subsection{Other model families}\label{sec:families}
\method{} carries over to EXAONE-4.0 and Granite-4.2, each steered by a small model of its own family, and shortens their CoT without losing accuracy (Figure~\ref{fig:tradeoff}b, Table~\ref{tab:families} in Appendix~\ref{app:families}).
The concise prompt, however, varies from model to model. It comes close to \method{} on Qwen3-32B, but on the other two \method{} removes about three times as much.
This depends on whether the model complies.
A prompt can only ask the large model to be brief, and models differ in how far they follow such a request, whereas \method{} asks the large model for nothing and changes its next-token distribution directly.

\subsection{Coding agent}\label{sec:agent}
We extend \method{} to a self-debugging coding agent on HumanEval+~\citep{chen2021humaneval,liu2024evalplus}, which writes a program, runs the tests and gets one repair turn on failure. \method{} is applied inside the CoT of every turn (Appendix~\ref{app:agent}).
Over the 164 problems, the agent thinks 20\% less per turn and improves its pass rate by 2.0 points over plain decoding (Table~\ref{tab:main}, Agent column).
DEER removes 41\% and improves the pass rate by 4.5 points. It ends the CoT once 80\% of the turn's thinking budget is spent, which leaves room for a program in every turn, whereas under plain decoding 16--20\% of the turns use the whole budget for thinking and write no program (Appendix~\ref{app:agent}).
NoWait and ThinkBrake remove more but lower the pass rate, ThinkBrake by 6.9 points.

\section{Analysis and discussion}\label{sec:analysis}
\subsection{What \method{} removes}\label{sec:where}
\begin{figure}[!t]
\casebox{Plain decoding}{3{,}414 thinking tokens, 2{,}369 after the first answer}{white}{%
\om{} \textbf{Therefore, the sum is 21 + 49 = 70.}\par
Wait, but let me \chk{check if there are any other divisors}.~\om\par
But let me \chk{check if there are other divisors}.~\om\par
Let me \chk{check 56's divisors again}:~\om\par
Is there a possibility that there are \chk{more divisors} if we consider other factorizations?~\om\par
But let me \chk{check if there's a possibility of other divisors}~\om\par
Therefore, answer is \textbf{70}.~\om}
\vspace{4pt}
\casebox{\method{} (0.6B)}{1{,}461 thinking tokens, 523 after the first answer}{oursbg}{%
\om{} \textbf{Their sum is 21 + 49 = 70.}\par
Wait, is that all? Let me \chk{verify if there are any other divisors}.~\om\par
But let me \chk{check with actual numbers} to make sure.~\om\par
But let me just make sure there isn't a \chk{negative divisor} or something?~\om\par
Therefore, the answer is $\boxed{70}$.~\om}
\caption{After the answer, \method{} checks briefly and stops, while plain decoding gets stuck repeating the same check. AIME 2025 problem 1, first seed: the CoT under each method from the first statement of the answer (bold) to its end. Checks are in red, and \om{} marks omitted text; full CoTs in Appendix~\ref{app:example}.}
\label{fig:example}
\end{figure}

\paragraph{Most removed tokens follow the first appearance of the answer.}
We split each run at the point where the correct answer first appears in the CoT (Appendix~\ref{app:where}).
\method{} cuts the thinking after this point by half and the thinking before it by a fifth (Figure~\ref{fig:removed}a), so most of the removed tokens lie after it on each of the six benchmarks on which the answer can be located (Table~\ref{tab:whereall}).
In Figure~\ref{fig:example}, both methods reach the answer after about a thousand tokens, and plain decoding then returns to the same check again and again.
Useful checks remain. On AIME 2025 problem 11, an algebraic check restores a wrongly excluded root, which is then included in the final answer (Appendix~\ref{app:usefulcheck}).

\begin{figure}[t]
\centering
\includegraphics[width=\columnwidth]{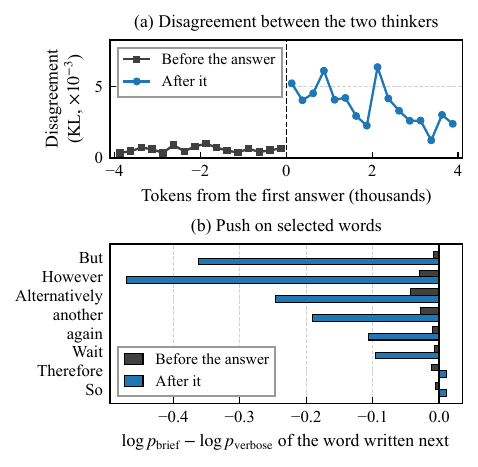}
\caption{The two thinkers disagree after the answer. Correct plain-decoding runs on AIME 2025, read by the small model under both instructions. (a) Divergence of the two next-token distributions, aligned at the first automatic match to the correct answer. (b) Brief minus verbose log-probability of selected words, before and after the answer.}
\label{fig:disagree}
\end{figure}

\paragraph{The two thinkers disagree after the answer.}
Read by the small model under both instructions, correct plain-decoding runs on AIME 2025 show a mean divergence between the two next-token distributions seven times larger after the answer than before it (Figure~\ref{fig:disagree}a), and the same holds under human-annotated answer boundaries (Appendix~\ref{app:where}).
After the answer the direction pushes down the words that open another round, such as \emph{But}, \emph{However} and \emph{Alternatively}, which it leaves almost untouched before the answer (Figure~\ref{fig:disagree}b).

\paragraph{The words of hesitation stay.}
\emph{Wait}, \emph{But} and \emph{Alternatively} occur at least as often per thousand thinking tokens under \method{} as under plain decoding, whereas NoWait suppresses them (Figure~\ref{fig:removed}b).
\method{} removes rounds of checking, and the checks that remain still begin with these words (Figure~\ref{fig:example}).

\subsection{\method{} complements training-based compression}\label{sec:laconic}
Training-based methods shorten the CoT by changing the weights, while \method{} acts only at decoding time, so the two can be combined.
We test this on a Qwen3-32B checkpoint fine-tuned with LACONIC~\citep{liu2026laconic}, which adds an adaptive length penalty to the task reward during reinforcement learning, and steer it with Qwen3-0.6B at $\alpha{=}5$.
Across mathematics, science and coding-agent benchmarks, \method{} further reduces thinking tokens by 29--52\% with unchanged task performance (Table~\ref{tab:laconic}).
\method{} thus complements training-based compression and applies to an already compressed model without further training.
\begin{table}[H]
\centering\footnotesize
\caption{\method{} further compresses a training-based compression model, Qwen3-32B fine-tuned with the adaptive length penalty of LACONIC (LACONIC-Qwen3-32B-3000). LACONIC: accuracy (\%) and mean thinking tokens of this model; +\method{}: change in points (95\% intervals) and in \%. Three seeds per problem. Agent: pass rate and thinking tokens per turn.}
\label{tab:laconic}
\setlength{\tabcolsep}{2pt}\renewcommand{\arraystretch}{1.05}
\resizebox{\columnwidth}{!}{%
\begin{tabular}{@{}ll rrr@{}}
\toprule
 & & Mathematics & Science & Agent \\
\cmidrule(lr){3-3}\cmidrule(lr){4-4}\cmidrule(l){5-5}
 & & \shortstack{Olympiad-\\Bench} & \shortstack{GPQA-\\Diamond} & \shortstack{HumanEval+\\agent} \\
\midrule
 & Acc. & 64.7 & 65.5 & 91.9 \\
\multirow{-2}{*}{LACONIC} & Tokens & 5,718 & 3,483 & 1,725 \\
\midrule
\rowcolor{oursbg} & Acc.\ change & $-1.3$ & $-1.7$ & $-1.2$ \\
\rowcolor{oursbg} & & \ci{-4.7}{+1.7} & \ci{-4.9}{+1.5} & \ci{-4.3}{+1.6} \\
\rowcolor{oursbg}\multirow{-3}{*}{\shortstack[l]{\textbf{+\method{}}\\\textbf{(0.6B)}}} & Token change & {\bfseries\boldmath $-42\%$} & {\bfseries\boldmath $-52\%$} & {\bfseries\boldmath $-29\%$} \\
\bottomrule
\end{tabular}}
\end{table}

\subsection{Inference cost}\label{sec:cost}
We measure end-to-end inference latency on a single B200 GPU, decoding one problem at a time on 14 problems, two from each of the seven benchmarks.
\method{} takes 119.4 seconds per problem on average, compared with 156.2 seconds for plain decoding and 137.5 seconds for the concise prompt, reducing latency by 24\% and 13\%, respectively.
The small model's two forward passes of 0.6B parameters per step add about 4\% to the FLOPs of the 32B model.

\section{Ablation study}\label{sec:ablation}
We vary the direction and the strength of the contrast and the size of the small model.

\subsection{Effect of the contrast direction}\label{sec:control}
The \emph{random-target} control tests whether the shortening comes from which tokens the small model pushes down or from perturbing the logits at all.
At every step it keeps the adjustment values $\alpha(\zc-\zv)$ but assigns them to randomly chosen tokens of the top-$p$ set, so the size and scope of the change stay the same and only its direction is lost.
With random targets, the thinking length stays within 8\% of plain decoding on every benchmark and changes by 1\% overall, against 31\% for \method{} (Table~\ref{tab:full}, Figure~\ref{fig:perproblem}).
The shortening therefore comes from the direction the small model supplies.

\subsection{Effect of the strength}\label{sec:strength}
\begin{figure}[t]
\centering
\includegraphics[width=\columnwidth]{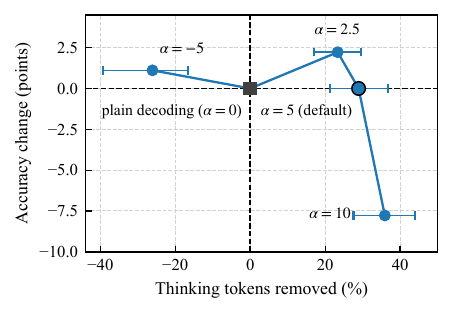}
\caption{Larger strengths remove more thinking tokens, and accuracy holds up to the default $\alpha{=}5$. AIME 2025, three seeds per problem, 95\% intervals (Table~\ref{tab:strength}).}
\label{fig:strength}
\end{figure}

On AIME 2025, \method{} removes more as the strength grows, with unchanged accuracy up to the default $\alpha{=}5$, and at $\alpha{=}10$ it loses 7.8 points (Figure~\ref{fig:strength}).
A negative strength, $\alpha{=}{-}5$, lengthens the CoT by 26\% instead, so the strength also sets the direction of the change.

\subsection{Effect of the small-model size}\label{sec:size}
We show that a larger small model at the same $\alpha$ acts as a stronger contrast. With Qwen3-1.7B or Qwen3-4B in place of Qwen3-0.6B, \method{} removes close to 60\% of the thinking tokens on AIME 2025 and, like $\alpha{=}10$ with the 0.6B model, goes past the point where accuracy holds (Table~\ref{tab:size}).
With Qwen3-32B itself as the small model (at $\alpha{=}1$), \method{} removes 26\%, against 29\% for the 0.6B model, so the smallest and cheapest model of the family, our default, gives up nothing.

\section{Related work}\label{sec:related}
\paragraph{Overthinking.}
Reasoning models spend a large share of their tokens re-deriving and re-verifying results they have already reached, even on simple problems~\citep{chen2024overthinking,sui2025stopoverthinking}, and their token spend tracks the difficulty of a problem only loosely~\citep{pu2025thoughtterminator}.
Section~\ref{sec:where} shows that the tokens \method{} removes come mainly from the checking after the answer.

\paragraph{Shortening the CoT at inference time.}
Methods that leave the model unchanged end the CoT once a budget is spent or the answer appears settled~\citep{muennighoff2025s1,yang2025deer,song2025thinkbrake,fu2024certaindex,pu2025thoughtterminator}, skip it~\citep{ma2025nothinking}, block the words that open another round of checking~\citep{wang2025nowait}, or ask for brevity in the prompt~\citep{renze2024ccot,han2024tale,xu2025chainofdraft}, which works only as far as the model follows the request~\citep{han2024tale,aggarwal2025l1}.
Activation steering adds a fixed vector to the hidden states, taken from concise and verbose CoTs~\citep{azizi2026asc,huang2025manifold} or from execution and reflection steps~\citep{chen2025seal}, with a strength that can follow a signal read from the model~\citep{lin2025thinkingspeed,li2026rebalance}. Budget Guidance~\citep{li2025budgetguidance} reweights the next-token distribution with a trained length predictor.
\method{} also reweights the next-token distribution, but its direction comes from an off-the-shelf small model, is recomputed from the context at every step and leaves the large model's own logit for \closetag{} untouched.

\paragraph{Training-based compression.}
Training-based methods add a length term or a token limit to reinforcement learning~\citep{aggarwal2025l1,hou2025thinkprune,luo2025o1pruner,liu2026laconic} or fine-tune on short traces~\citep{chen2024overthinking}.
Each model needs its own training run, and the trade-off between length and accuracy is fixed once training ends.
\method{} leaves the model unchanged, sets the trade-off with one strength at decoding time and can be added to a model trained this way (Section~\ref{sec:laconic}).

\paragraph{Contrastive decoding.}
Contrastive decoding scores the next token by the difference between an expert and a weaker amateur model~\citep{li2023contrastive,obrien2023cdreasoning}, and DExperts and proxy-tuning steer a large model with small models that share only its vocabulary~\citep{liu2021dexperts,liu2024proxytuning}.
ThinkLogit~\citep{zhang2025logitarithmetic} and ProxyThinker~\citep{xiao2025proxythinker} add the difference between a small reasoning model and its base to make a large model reason more, Thinking by Subtraction~\citep{tang2026subtraction} contrasts a model with a masked copy of its own context, and PromptCD~\citep{bi2026promptcd} and \citet{dong2026steer} contrast two prompts on the steered model itself.
\method{} draws the contrast from an off-the-shelf small model under two instructions and uses it to shorten the CoT.
Small models also work beside large ones as drafters in speculative decoding~\citep{leviathan2023speculative}, as token suggesters in Nudging~\citep{fei2025nudging} and as sentence completers in FoReaL decoding~\citep{li2025foreal}.

\FloatBarrier   %
\section{Conclusion}
\method{} adds the difference between a small reasoning model's brief and verbose thinkers to a large reasoning model's logits while it writes its CoT.
On seven benchmarks it removes 31\% of the thinking tokens of Qwen3-32B with unchanged accuracy, more than five inference-time baselines.
It also shortens the CoT of a coding agent, and on two other model families it removes about three times as much as a concise prompt.
Combined with training-based compression, it further shortens the CoT of a model already trained for brevity.
Our analyses support the explanation that \method{} reduces redundant checking while retaining useful verification.

\bibliography{custom}

\begin{thebibliography}{51}
\providecommand{\natexlab}[1]{#1}

\bibitem[{Aggarwal and Welleck(2025)}]{aggarwal2025l1}
Pranjal Aggarwal and Sean Welleck. 2025.
\newblock {L1}: Controlling how long a reasoning model thinks with
  reinforcement learning.
\newblock In \emph{Second Conference on Language Modeling (COLM)}.
\newblock ArXiv:2503.04697.

\bibitem[{Azizi et~al.(2026)Azizi, Potraghloo, Kundu, and
  Pedram}]{azizi2026asc}
Seyedarmin Azizi, Erfan~Baghaei Potraghloo, Souvik Kundu, and Massoud Pedram.
  2026.
\newblock \href {https://doi.org/10.18653/v1/2026.findings-acl.1828}
  {Activation steering for chain-of-thought compression}.
\newblock In \emph{Findings of the Association for Computational Linguistics:
  ACL 2026}, pages 36676--36687.

\bibitem[{Bi et~al.(2026)Bi, Ge, Liu, He, Tong, Chen, Mei, Li, Wang, Cai, Yang,
  and Cheng}]{bi2026promptcd}
Baolong Bi, Yuyao Ge, Shenghua Liu, Yuchen He, Siqian Tong, Lizhe Chen, Lingrui
  Mei, Zehao Li, Yiwei Wang, Yujun Cai, Ming-Hsuan Yang, and Xueqi Cheng. 2026.
\newblock {PromptCD}: Test-time behavior enhancement via polarity-prompt
  contrastive decoding.
\newblock \emph{arXiv preprint arXiv:2602.20696}.

\bibitem[{Chen et~al.(2021)Chen, Tworek, Jun, Yuan, {Ponde de Oliveira Pinto},
  Kaplan, Edwards, Burda, Joseph, Brockman et~al.}]{chen2021humaneval}
Mark Chen, Jerry Tworek, Heewoo Jun, Qiming Yuan, Henrique {Ponde de Oliveira
  Pinto}, Jared Kaplan, Harri Edwards, Yuri Burda, Nicholas Joseph, Greg
  Brockman, et~al. 2021.
\newblock \href {https://arxiv.org/abs/2107.03374} {Evaluating large language
  models trained on code}.
\newblock \emph{arXiv preprint arXiv:2107.03374}.

\bibitem[{Chen et~al.(2025)Chen, Zhang, Hong, Kundu, and Wang}]{chen2025seal}
Runjin Chen, Zhenyu Zhang, Junyuan Hong, Souvik Kundu, and Zhangyang Wang.
  2025.
\newblock \href {https://openreview.net/forum?id=klPszYDIRT} {{SEAL}: Steerable
  reasoning calibration of large language models for free}.
\newblock In \emph{Second Conference on Language Modeling (COLM)}.

\bibitem[{Chen et~al.(2024)Chen, Xu, Liang, He, Pang, Yu, Song, Liu, Zhou,
  Zhang, Wang, Tu, Mi, and Yu}]{chen2024overthinking}
Xingyu Chen, Jiahao Xu, Tian Liang, Zhiwei He, Jianhui Pang, Dian Yu, Linfeng
  Song, Qiuzhi Liu, Mengfei Zhou, Zhuosheng Zhang, Rui Wang, Zhaopeng Tu,
  Haitao Mi, and Dong Yu. 2024.
\newblock Do {NOT} think that much for 2+3=? on the overthinking of o1-like
  {LLMs}.
\newblock \emph{arXiv preprint arXiv:2412.21187}.

\bibitem[{Dong et~al.(2026)Dong, Hu, Hui, and Collier}]{dong2026steer}
Yijiang~River Dong, Tiancheng Hu, Zheng Hui, and Nigel Collier. 2026.
\newblock Steer model beyond assistant: Controlling system prompt strength via
  contrastive decoding.
\newblock \emph{arXiv preprint arXiv:2601.06403}.

\bibitem[{Fei et~al.(2025)Fei, Razeghi, and Singh}]{fei2025nudging}
Yu~Fei, Yasaman Razeghi, and Sameer Singh. 2025.
\newblock Nudging: Inference-time alignment of {LLMs} via guided decoding.
\newblock In \emph{Proceedings of the 63rd Annual Meeting of the Association
  for Computational Linguistics (ACL)}.
\newblock ArXiv:2410.09300.

\bibitem[{Fu et~al.(2024)Fu, Chen, Zhu, Fu, Dai, Zhuang, Ma, Qiao, Rosing,
  Stoica, and Zhang}]{fu2024certaindex}
Yichao Fu, Junda Chen, Siqi Zhu, Zheyu Fu, Zhongdongming Dai, Yonghao Zhuang,
  Yian Ma, Aurick Qiao, Tajana Rosing, Ion Stoica, and Hao Zhang. 2024.
\newblock Efficiently scaling {LLM} reasoning with {Certaindex}.
\newblock \emph{arXiv preprint arXiv:2412.20993}.

\bibitem[{Han et~al.(2025)Han, Wang, Fang, Zhao, Ma, and Chen}]{han2024tale}
Tingxu Han, Zhenting Wang, Chunrong Fang, Shiyu Zhao, Shiqing Ma, and Zhenyu
  Chen. 2025.
\newblock \href {https://doi.org/10.18653/v1/2025.findings-acl.1274}
  {Token-budget-aware {LLM} reasoning}.
\newblock In \emph{Findings of the Association for Computational Linguistics:
  ACL 2025}, pages 24842--24855.

\bibitem[{He et~al.(2024)He, Luo, Bai, Hu, Thai, Shen, Hu, Han, Huang, Zhang,
  Liu, Qi, Liu, and Sun}]{he2024olympiadbench}
Chaoqun He, Renjie Luo, Yuzhuo Bai, Shengding Hu, Zhen Thai, Junhao Shen, Jinyi
  Hu, Xu~Han, Yujie Huang, Yuxiang Zhang, Jie Liu, Lei Qi, Zhiyuan Liu, and
  Maosong Sun. 2024.
\newblock \href {https://doi.org/10.18653/v1/2024.acl-long.211}
  {{O}lympiad{B}ench: A challenging benchmark for promoting {AGI} with
  olympiad-level bilingual multimodal scientific problems}.
\newblock In \emph{Proceedings of the 62nd Annual Meeting of the Association
  for Computational Linguistics (Volume 1: Long Papers)}, pages 3828--3850,
  Bangkok, Thailand. Association for Computational Linguistics.

\bibitem[{Hendrycks et~al.(2021)Hendrycks, Burns, Kadavath, Arora, Basart,
  Tang, Song, and Steinhardt}]{hendrycks2021math}
Dan Hendrycks, Collin Burns, Saurav Kadavath, Akul Arora, Steven Basart, Eric
  Tang, Dawn Song, and Jacob Steinhardt. 2021.
\newblock Measuring mathematical problem solving with the {MATH} dataset.
\newblock In \emph{Proceedings of the Neural Information Processing Systems
  Track on Datasets and Benchmarks}.
\newblock ArXiv:2103.03874.

\bibitem[{{HMMT}(2025)}]{hmmt2025}
{HMMT}. 2025.
\newblock \href {https://www.hmmt.org/www/archive/282} {{HMMT February 2025}}.
\newblock \url{https://www.hmmt.org/www/archive/282}.

\bibitem[{Holtzman et~al.(2020)Holtzman, Buys, Du, Forbes, and
  Choi}]{holtzman2020nucleus}
Ari Holtzman, Jan Buys, Li~Du, Maxwell Forbes, and Yejin Choi. 2020.
\newblock The curious case of neural text degeneration.
\newblock \emph{Proceedings of ICLR}.

\bibitem[{Hou et~al.(2025)Hou, Zhang, Ji, Liu, Qian, Andreas, and
  Chang}]{hou2025thinkprune}
Bairu Hou, Yang Zhang, Jiabao Ji, Yujian Liu, Kaizhi Qian, Jacob Andreas, and
  Shiyu Chang. 2025.
\newblock {ThinkPrune}: Pruning long chain-of-thought of {LLMs} via
  reinforcement learning.
\newblock \emph{arXiv preprint arXiv:2504.01296}.

\bibitem[{Huang et~al.(2025)Huang, Chen, Ruan, Zhang, Wei, and
  Dong}]{huang2025manifold}
Yao Huang, Huanran Chen, Shouwei Ruan, Yichi Zhang, Xingxing Wei, and Yinpeng
  Dong. 2025.
\newblock \href {https://openreview.net/forum?id=49Rc51iCso} {Mitigating
  overthinking in large reasoning models via manifold steering}.
\newblock In \emph{Advances in Neural Information Processing Systems}.

\bibitem[{{IBM Granite Team}(2026)}]{ibm2026granite42}
{IBM Granite Team}. 2026.
\newblock \href {https://huggingface.co/ibm-granite/granite-4.2-30b} {{Granite}
  4.2 language models}.
\newblock \url{https://huggingface.co/ibm-granite/granite-4.2-30b}.

\bibitem[{Jain et~al.(2025)Jain, Han, Gu, Li, Yan, Zhang, Wang, Solar-Lezama,
  Sen, and Stoica}]{jain2025livecodebench}
Naman Jain, King Han, Alex Gu, Wen-Ding Li, Fanjia Yan, Tianjun Zhang, Sida~I.
  Wang, Armando Solar-Lezama, Koushik Sen, and Ion Stoica. 2025.
\newblock \href
  {https://proceedings.iclr.cc/paper_files/paper/2025/hash/94074dd5a072d28ff75a76dabed43767-Abstract-Conference.html}
  {{LiveCodeBench}: Holistic and contamination free evaluation of large
  language models for code}.
\newblock In \emph{International Conference on Learning Representations}, pages
  58791--58831.

\bibitem[{Kwon et~al.(2023)Kwon, Li, Zhuang, Sheng, Zheng, Yu, Gonzalez, Zhang,
  and Stoica}]{kwon2023vllm}
Woosuk Kwon, Zhuohan Li, Siyuan Zhuang, Ying Sheng, Lianmin Zheng, Cody~Hao Yu,
  Joseph~E. Gonzalez, Hao Zhang, and Ion Stoica. 2023.
\newblock \href {https://doi.org/10.1145/3600006.3613165} {Efficient memory
  management for large language model serving with {PagedAttention}}.
\newblock In \emph{Proceedings of the 29th Symposium on Operating Systems
  Principles}, pages 611--626, New York, NY, USA. ACM.

\bibitem[{Kydl{\'\i}{\v{c}}ek(2025)}]{mathverify}
Hynek Kydl{\'\i}{\v{c}}ek. 2025.
\newblock \href {https://github.com/huggingface/Math-Verify} {{Math-Verify}:
  Math verification library}.
\newblock \url{https://github.com/huggingface/Math-Verify}.

\bibitem[{Leviathan et~al.(2023)Leviathan, Kalman, and
  Matias}]{leviathan2023speculative}
Yaniv Leviathan, Matan Kalman, and Yossi Matias. 2023.
\newblock Fast inference from transformers via speculative decoding.
\newblock In \emph{Proceedings of the 40th International Conference on Machine
  Learning (ICML)}.

\bibitem[{{LG AI Research}(2025)}]{lgai2025exaone4}
{LG AI Research}. 2025.
\newblock \href {https://arxiv.org/abs/2507.11407} {{EXAONE} 4.0: Unified large
  language models integrating non-reasoning and reasoning modes}.
\newblock \emph{arXiv preprint arXiv:2507.11407}.

\bibitem[{Li et~al.(2026{\natexlab{a}})Li, Zhao, Zhang, and
  Gan}]{li2025budgetguidance}
Junyan Li, Wenshuo Zhao, Yang Zhang, and Chuang Gan. 2026{\natexlab{a}}.
\newblock \href {https://doi.org/10.18653/v1/2026.findings-acl.1866} {Steering
  {LLM} thinking with budget guidance}.
\newblock In \emph{Findings of the Association for Computational Linguistics:
  ACL 2026}, pages 37437--37451.

\bibitem[{Li et~al.(2025)Li, Yang, Wang, Li, Lin, Zhou, and
  Wang}]{li2025foreal}
Ming Li, Zhengyuan Yang, Xiyao Wang, Dianqi Li, Kevin Lin, Tianyi Zhou, and
  Lijuan Wang. 2025.
\newblock What makes reasoning models different? {Follow} the reasoning leader
  for efficient decoding.
\newblock \emph{arXiv preprint arXiv:2506.06998}.

\bibitem[{Li et~al.(2023)Li, Holtzman, Fried, Liang, Eisner, Hashimoto,
  Zettlemoyer, and Lewis}]{li2023contrastive}
Xiang~Lisa Li, Ari Holtzman, Daniel Fried, Percy Liang, Jason Eisner, Tatsunori
  Hashimoto, Luke Zettlemoyer, and Mike Lewis. 2023.
\newblock Contrastive decoding: Open-ended text generation as optimization.
\newblock In \emph{Proceedings of the 61st Annual Meeting of the Association
  for Computational Linguistics (ACL)}.

\bibitem[{Li et~al.(2026{\natexlab{b}})Li, Tu, Ding, Wang, Zhen, Chen, Li, and
  Tian}]{li2026rebalance}
Yulin Li, Tengyao Tu, Li~Ding, Junjie Wang, Hui-Ling Zhen, Yixin Chen, Yong Li,
  and Zhuotao Tian. 2026{\natexlab{b}}.
\newblock \href {https://openreview.net/forum?id=cJseWJJ5IM} {Efficient
  reasoning with balanced thinking}.
\newblock In \emph{International Conference on Learning Representations}.

\bibitem[{Lightman et~al.(2024)Lightman, Kosaraju, Burda, Edwards, Baker, Lee,
  Leike, Schulman, Sutskever, and Cobbe}]{lightman2024verify}
Hunter Lightman, Vineet Kosaraju, Yura Burda, Harri Edwards, Bowen Baker, Teddy
  Lee, Jan Leike, John Schulman, Ilya Sutskever, and Karl Cobbe. 2024.
\newblock Let's verify step by step.
\newblock In \emph{The Twelfth International Conference on Learning
  Representations (ICLR)}.
\newblock ArXiv:2305.20050.

\bibitem[{Lin et~al.(2025)Lin, Fu, Chen, Chen, Xie, Wang, Cai, Wang, and
  Ye}]{lin2025thinkingspeed}
Zhengkai Lin, Zhihang Fu, Ze~Chen, Chao Chen, Liang Xie, Wenxiao Wang, Deng
  Cai, Zheng Wang, and Jieping Ye. 2025.
\newblock \href {https://openreview.net/forum?id=ESELaMThLN} {Controlling
  thinking speed in reasoning models}.
\newblock In \emph{Advances in Neural Information Processing Systems}.

\bibitem[{Liu et~al.(2024)Liu, Han, Wang, Tsvetkov, Choi, and
  Smith}]{liu2024proxytuning}
Alisa Liu, Xiaochuang Han, Yizhong Wang, Yulia Tsvetkov, Yejin Choi, and
  Noah~A. Smith. 2024.
\newblock Tuning language models by proxy.
\newblock In \emph{First Conference on Language Modeling (COLM)}.

\bibitem[{Liu et~al.(2021)Liu, Sap, Lu, Swayamdipta, Bhagavatula, Smith, and
  Choi}]{liu2021dexperts}
Alisa Liu, Maarten Sap, Ximing Lu, Swabha Swayamdipta, Chandra Bhagavatula,
  Noah~A. Smith, and Yejin Choi. 2021.
\newblock {DExperts}: Decoding-time controlled text generation with experts and
  anti-experts.
\newblock In \emph{Proceedings of the 59th Annual Meeting of the Association
  for Computational Linguistics (ACL)}.

\bibitem[{Liu et~al.(2026)Liu, Zhao, Liu, Ye, Szepesv{\'a}ri, and
  Yang}]{liu2026laconic}
Chang Liu, Yiran Zhao, Lawrence Liu, Yaoqi Ye, Csaba Szepesv{\'a}ri, and Lin~F.
  Yang. 2026.
\newblock \href {https://doi.org/10.48550/arXiv.2602.14468} {{LACONIC}:
  Length-aware constrained reinforcement learning for {LLM}}.
\newblock \emph{arXiv preprint arXiv:2602.14468}.

\bibitem[{Liu et~al.(2023)Liu, Xia, Wang, and Zhang}]{liu2024evalplus}
Jiawei Liu, Chunqiu~Steven Xia, Yuyao Wang, and Lingming Zhang. 2023.
\newblock Is your code generated by {ChatGPT} really correct? rigorous
  evaluation of large language models for code generation.
\newblock In \emph{Advances in Neural Information Processing Systems},
  volume~36.

\bibitem[{Luo et~al.(2025)Luo, Shen, He, Wang, Liu, Li, Tan, Cao, and
  Tao}]{luo2025o1pruner}
Haotian Luo, Li~Shen, Haiying He, Yibo Wang, Shiwei Liu, Wei Li, Naiqiang Tan,
  Xiaochun Cao, and Dacheng Tao. 2025.
\newblock {O1-Pruner}: Length-harmonizing fine-tuning for {O1}-like reasoning
  pruning.
\newblock \emph{arXiv preprint arXiv:2501.12570}.

\bibitem[{Ma et~al.(2025)Ma, He, Snell, Griggs, Min, and
  Zaharia}]{ma2025nothinking}
Wenjie Ma, Jingxuan He, Charlie Snell, Tyler Griggs, Sewon Min, and Matei
  Zaharia. 2025.
\newblock Reasoning models can be effective without thinking.
\newblock \emph{arXiv preprint arXiv:2504.09858}.

\bibitem[{{Mathematical Association of America}(2023)}]{maa_amc}
{Mathematical Association of America}. 2023.
\newblock \href {https://maa.org/student-programs/amc/} {{American Mathematics
  Competitions (AMC 12)}}.
\newblock \url{https://maa.org/student-programs/amc/}.
\newblock AMC 12A and 12B, 2023.

\bibitem[{{Mathematical Association of America}(2025)}]{maa_aime}
{Mathematical Association of America}. 2025.
\newblock \href {https://maa.org/maa-invitational-competitions/} {{American
  Invitational Mathematics Examination (AIME)}}.
\newblock \url{https://maa.org/maa-invitational-competitions/}.

\bibitem[{{Mathematical Association of America}(2026)}]{maa_aime26}
{Mathematical Association of America}. 2026.
\newblock \href {https://maa.org/maa-invitational-competitions/} {{American
  Invitational Mathematics Examination (AIME) 2026}}.
\newblock \url{https://maa.org/maa-invitational-competitions/}.

\bibitem[{Muennighoff et~al.(2025)Muennighoff, Yang, Shi, Li, Fei-Fei,
  Hajishirzi, Zettlemoyer, Liang, Cand{\`e}s, and
  Hashimoto}]{muennighoff2025s1}
Niklas Muennighoff, Zitong Yang, Weijia Shi, Xiang~Lisa Li, Li~Fei-Fei,
  Hannaneh Hajishirzi, Luke Zettlemoyer, Percy Liang, Emmanuel Cand{\`e}s, and
  Tatsunori Hashimoto. 2025.
\newblock s1: Simple test-time scaling.
\newblock \emph{arXiv preprint arXiv:2501.19393}.

\bibitem[{O'Brien and Lewis(2023)}]{obrien2023cdreasoning}
Sean O'Brien and Mike Lewis. 2023.
\newblock Contrastive decoding improves reasoning in large language models.
\newblock \emph{arXiv preprint arXiv:2309.09117}.

\bibitem[{Pu et~al.(2025)Pu, Saxon, Hua, and Wang}]{pu2025thoughtterminator}
Xiao Pu, Michael Saxon, Wenyue Hua, and William~Yang Wang. 2025.
\newblock {ThoughtTerminator}: Benchmarking, calibrating, and mitigating
  overthinking in reasoning models.
\newblock \emph{arXiv preprint arXiv:2504.13367}.

\bibitem[{Rein et~al.(2024)Rein, Hou, Stickland, Petty, Pang, Dirani, Michael,
  and Bowman}]{rein2024gpqa}
David Rein, Betty~Li Hou, Asa~Cooper Stickland, Jackson Petty, Richard~Yuanzhe
  Pang, Julien Dirani, Julian Michael, and Samuel~R. Bowman. 2024.
\newblock \href {https://openreview.net/forum?id=Ti67584b98} {{GPQA}: A
  graduate-level {Google}-proof {Q\&A} benchmark}.
\newblock In \emph{First Conference on Language Modeling}.

\bibitem[{Renze and Guven(2024)}]{renze2024ccot}
Matthew Renze and Erhan Guven. 2024.
\newblock The benefits of a concise chain of thought on problem-solving in
  large language models.
\newblock In \emph{2nd International Conference on Foundation and Large
  Language Models (FLLM)}.
\newblock ArXiv:2401.05618.

\bibitem[{Song et~al.(2026)Song, Oh, Lee, Jo, and Jo}]{song2025thinkbrake}
Sangjun Song, Minjae Oh, Seungkyu Lee, Sungmin Jo, and Yohan Jo. 2026.
\newblock \href {https://doi.org/10.18653/v1/2026.findings-acl.1095}
  {{ThinkBrake}: Efficient reasoning via log-probability margin guided
  decoding}.
\newblock In \emph{Findings of the Association for Computational Linguistics:
  ACL 2026}, pages 21765--21790.

\bibitem[{Sui et~al.(2025)Sui, Chuang, Wang, Zhang, Zhang, Yuan, Liu, Wen,
  Zhong, Zou, Chen, and Hu}]{sui2025stopoverthinking}
Yang Sui, Yu-Neng Chuang, Guanchu Wang, Jiamu Zhang, Tianyi Zhang, Jiayi Yuan,
  Hongyi Liu, Andrew Wen, Shaochen Zhong, Na~Zou, Hanjie Chen, and Xia Hu.
  2025.
\newblock Stop overthinking: A survey on efficient reasoning for large language
  models.
\newblock \emph{Transactions on Machine Learning Research}.
\newblock ArXiv:2503.16419.

\bibitem[{Tang et~al.(2026)Tang, Gao, Zhao, Ma, Jin, Yang, and
  Zou}]{tang2026subtraction}
Lexiang Tang, Weihao Gao, Bingchen Zhao, Lu~Ma, Qiao Jin, Bang Yang, and
  Yuexian Zou. 2026.
\newblock Thinking by subtraction: Confidence-driven contrastive decoding for
  {LLM} reasoning.
\newblock \emph{arXiv preprint arXiv:2602.18232}.

\bibitem[{Wang et~al.(2025)Wang, Feng, Chen, Chu, Krishna, and
  Zhou}]{wang2025nowait}
Chenlong Wang, Yuanning Feng, Dongping Chen, Zhaoyang Chu, Ranjay Krishna, and
  Tianyi Zhou. 2025.
\newblock Wait, we don't need to ``wait''! removing thinking tokens improves
  reasoning efficiency.
\newblock \emph{arXiv preprint arXiv:2506.08343}.

\bibitem[{Xiao et~al.(2025)Xiao, Koo, Ouyang, Hernandez, Meng, and
  Ordonez}]{xiao2025proxythinker}
Zilin Xiao, Jaywon Koo, Siru Ouyang, Jefferson Hernandez, Yu~Meng, and Vicente
  Ordonez. 2025.
\newblock {ProxyThinker}: Test-time guidance through small visual reasoners.
\newblock \emph{arXiv preprint arXiv:2505.24872}.

\bibitem[{Xu et~al.(2025)Xu, Xie, Zhao, and He}]{xu2025chainofdraft}
Silei Xu, Wenhao Xie, Lingxiao Zhao, and Pengcheng He. 2025.
\newblock Chain of draft: Thinking faster by writing less.
\newblock \emph{arXiv preprint arXiv:2502.18600}.

\bibitem[{Yang et~al.(2025{\natexlab{a}})Yang, Li, Yang et~al.}]{yang2025qwen3}
An~Yang, Anfeng Li, Baosong Yang, et~al. 2025{\natexlab{a}}.
\newblock {Qwen3} technical report.
\newblock \emph{arXiv preprint arXiv:2505.09388}.

\bibitem[{Yang et~al.(2025{\natexlab{b}})Yang, Si, Duan, Zhu, Zhu, Li, Chen,
  Lin, and Wang}]{yang2025deer}
Chenxu Yang, Qingyi Si, Yongjie Duan, Zheliang Zhu, Chenyu Zhu, Qiaowei Li,
  Minghui Chen, Zheng Lin, and Weiping Wang. 2025{\natexlab{b}}.
\newblock Dynamic early exit in reasoning models.
\newblock \emph{arXiv preprint arXiv:2504.15895}.

\bibitem[{Zhang et~al.(2026)Zhang, Khalifa, Zhang, Liu, Lee, Zhang,
  Fatahi~Bayat, and Wang}]{zhang2025logitarithmetic}
Yunxiang Zhang, Muhammad Khalifa, Lechen Zhang, Xin Liu, Ayoung Lee,
  Xinliang~Frederick Zhang, Farima Fatahi~Bayat, and Lu~Wang. 2026.
\newblock Logit arithmetic elicits long reasoning capabilities without
  training.
\newblock In \emph{Findings of the Association for Computational Linguistics:
  ACL 2026}.
\newblock ArXiv:2510.09354.

\end{thebibliography}

\appendix
\setlength{\intextsep}{6pt plus 2pt minus 2pt}\setlength{\textfloatsep}{10pt plus 2pt minus 2pt}   %
\makeatletter\setlength{\@fptop}{0pt}\setlength{\@dblfptop}{0pt}\makeatother   %
\raggedbottom   %
\FloatBarrier
\section{Algorithm and random-target control}\label{app:setup}
\begin{algorithm}[H]
\caption{One decoding step inside the CoT}
\label{alg:step}
\small
\begin{algorithmic}[1]
\Require prompt $x$, tokens $y_{<t}$, strength $\alpha$, top-$p$; sampler $(T, p, k)$
\State $\zb \gets M(x, y_{<t})$
\State $\zc \gets D(x, y_{<t};\text{brief})$;\quad $\zv \gets D(x, y_{<t};\text{verbose})$
\State $S \gets \mathcal{V}_p(\mathrm{softmax}(\zb))$ \Comment{top-$p$ set at temperature one}
\State $g \gets \zc-\zv$
\State $z \gets \zb+\alpha\,g$
\State $z[c] \gets \zb[c]$ for $c \in \{\closetag{},\,\mathrm{EOS}\}$
\State $z[v]\gets-\infty$ for $v\notin S$ \Comment{the mask is applied last}
\State \Return a token sampled from $z$ with $(T, p, k)$, appended to all three streams
\end{algorithmic}
\end{algorithm}

\paragraph{Where the contrast applies.}
The contrast is added only while the large model writes its CoT, and the answer after \closetag{} is decoded by the large model alone.
Inside the CoT, the algorithm restores the large model's own logits for \closetag{} and the end-of-sequence token, so the CoT ends only where the large model itself would consider ending it.
At such a point the contrast changes the competition. By lowering the tokens that open another round of checking, it lets the large model's own \closetag{} win more often.
The mask is applied last, so the candidates are the top-$p$ set of the large model's distribution at temperature one and the contrast never brings in a token outside it.
The method has one hyperparameter, the strength $\alpha$.

\paragraph{Random-target control.}
The control differs from Algorithm~\ref{alg:step} in one step. After $g$ is formed, it draws a uniformly random permutation of the positions in the top-$p$ set $S$ and assigns the entries of $g$ at those positions to the permuted positions. Entries outside $S$ are masked anyway.
The permutation is seeded by the run's seed and the step index.
Everything else, including $\alpha$, the restoration of the two logits and the mask, is identical to \method{}.

\FloatBarrier
\section{Prompts}\label{app:prompts}
Every stream receives the problem followed by one instruction line.
On GPQA-Diamond the answer is the letter of the chosen option, and on LiveCodeBench the instruction asks for a program in a code block.
The brief and the verbose thinker append one sentence each, and the large model's own stream carries neither.
The concise-prompt setting gives the large model the brief sentence.

\promptbox{Instruction (all streams)}{Please reason step by step, and put your final answer within \texttt{\textbackslash boxed\{\}}.}
\promptbox{Brief thinker (appended)}{Be concise: go straight to the answer with the fewest steps. Once you have a result, do NOT re-derive, re-verify, or repeat any calculation, and do not second-guess it.}
\promptbox{Verbose thinker (appended)}{Be extremely thorough: double-check every step, re-verify each calculation in more than one way, and exhaustively consider alternative cases before you commit to a final answer.}
\promptbox{NoWait: blocked words}{wait, alternatively, hmm, but, however, alternative, another, check, double-check, oh, maybe, verify, other, again, now, ah, any. Each word is blocked in every single-token surface form (capitalized, with a leading space, after a full stop, in capitals), 89 token ids in total, with a logit bias of $-100$ throughout the generation. The seventeen words are those of \citet{wang2025nowait}.}

\FloatBarrier
\section{The logit difference as a ratio of probabilities}\label{app:ratio}
Write $p_M$, $p_{\mathrm{b}}$ and $p_{\mathrm{v}}$ for the softmax distributions of $\zb$, $\zc$ and $\zv$, and $Z_M$, $Z_{\mathrm{b}}$, $Z_{\mathrm{v}}$ for their normalizers.
For any token $w$, $\zb[w]=\log p_M(w)+\log Z_M$ and likewise for the two thinkers. The sampler divides $z$ by the temperature $T$ before the softmax, so
\begin{align*}
&\exp\bigl(\bigl(\zb[w]+\alpha(\zc[w]-\zv[w])\bigr)/T\bigr)\\
&\qquad=\;Z_M^{1/T}\Bigl(\tfrac{Z_{\mathrm{b}}}{Z_{\mathrm{v}}}\Bigr)^{\alpha/T}\,p_M(w)^{1/T}\Bigl[\tfrac{p_{\mathrm{b}}(w)}{p_{\mathrm{v}}(w)}\Bigr]^{\alpha/T}.
\end{align*}
The factor in front does not depend on $w$ and cancels in the softmax, and $p_M(w)^{1/T}$ normalized is $p_M^{T}$, the distribution plain decoding samples from, which gives Equation~\eqref{eq:poe}. At $T{=}1$ the exponent is $\alpha$.
The top-$p$ mask acts on $z$ after this step, so it applies to the reweighted distribution.
Restoring the \closetag{} and end-of-sequence logits after this step also removes the factor $(Z_{\mathrm{b}}/Z_{\mathrm{v}})^{\alpha/T}$ from these two tokens alone, so against every other candidate their odds are multiplied by $e^{-\alpha\Delta/T}$, where $\Delta=\log Z_{\mathrm{b}}-\log Z_{\mathrm{v}}$.
On the plain-decoding CoTs of the first six problems of AIME 2025 (one run each, 42{,}076 positions), $\Delta$ has mean $-0.04$ and standard deviation $0.11$, so at $\alpha/T{=}8.3$ this factor is $1.33$ at the median and lies between $0.54$ and $3.58$ on nine positions in ten.
The random-target control permutes the values of $\zc-\zv$, each of which contains $\Delta$, within the top-$p$ set and restores the same two logits, so it carries the same factor.
It changes the thinking length by 1\% over the seven benchmarks (Table~\ref{tab:full}), so the shortening under \method{} comes from which tokens the small model pushes down.

\FloatBarrier
\section{Implementation details}\label{app:impl}
The large model runs in vLLM 0.29~\citep{kwon2023vllm} with tensor parallelism. We load the small model into the same vLLM model as the large one, so that a single request drives all three streams of Figure~\ref{fig:schematic}.
The small model keeps one copy of its linear weights but two copies of its attention layers, one for the brief thinker and one for the verbose thinker. The two thinkers go through the linear layers together, as a doubled batch, and through attention separately. Under tensor parallelism the small model is split across the GPUs by attention head, like the large model.
The three streams differ only in the instruction sentence, which sits between the problem and the last few tokens of the chat template (Appendix~\ref{app:prompts}). Everything else, the prompt, those template tokens and the generated tokens, uses the same slot mapping in the three streams' paged KV caches, and we keep it in one paged key--value cache. The request's block table is allocated for the large model's sequence, and the small model's attention layers, which have the same cache layout (eight key--value heads of dimension 128 in bfloat16, split across the GPUs), write their keys and values into the same blocks. vLLM then allocates, accounts for and frees the three streams as one, with no padding and no duplicated copies.
The keys and values of the instruction sentence itself, at most 64 tokens per thinker, live in a small buffer per request that is allocated together with the model. The position IDs are those of each stream's own sequence, so the thinkers see the template tokens and the generated tokens shifted by the length of their sentence. A key stores its own position, so the order of the blocks does not have to match the order of the positions.
A thinker's attention is computed in two parts, paged attention over the request's blocks and dense attention over the instruction buffer, which are merged by their log-sum-exp (vLLM's cascade-attention kernel does exactly this). The result is the attention over the full sequence.
In the first step all three streams prefill the problem. The small model then reads its two instruction sentences against the cached problem and fills the buffer, after which decoding proceeds with every sampled token fed to all three streams.
The forward pass returns the three hidden states. At the logits step we apply the three output heads, form $\zb+\alpha(\zc-\zv)$ for the requests that are still inside their CoT, restore the \closetag{} and end-of-sequence logits, mask to the large model's top-$p$ set (the random-target control permutes the difference within that set here) and pass $z$ to vLLM's sampler, which draws with the request's temperature, top-$p$, top-$k$ and seed.
Whether a request has left its CoT is tracked on the GPU from the tokens fed to the model, so asynchronous scheduling and CUDA graphs work as usual. After \closetag{} the contrast is no longer applied.
The small model runs on a second CUDA stream and overlaps the large model's forward pass, and one CUDA graph per decoding step covers both models.
With one problem in flight, a decoding step is bound by reading the large model's weights (65.5 GB in bfloat16), and the small model's kernels run on the second stream while those weights stream, with vLLM's fused kernels for its normalization and rotary embedding and a single call of its output head for both thinkers.
We time each tokenized request from submission to the engine until the complete response text is returned, including prefill and decoding.
Before a run we check that the two vocabularies have the same size (151{,}936 for Qwen3) and that the instruction, \thinktag{} and \closetag{} encode to the same ids with both tokenizers.
\paragraph{NoWait.}
The blocked words are the seventeen of \citet{wang2025nowait}, listed in Appendix~\ref{app:prompts}.
The paper expands each word to every vocabulary token that contains it as a substring and then removes unreasonable matches by hand (its example is \emph{Ohio} for \emph{oh}). On the Qwen3 vocabulary the substring rule matches 918 tokens, most of them unrelated words such as \emph{know}, \emph{many}, \emph{company} and \emph{button}, and the filtered list is not published.
We block the 89 single-token forms of the words: the word itself, with a leading space, capitalized, in upper case and after a period, which are the forms the paper gives as its example for \emph{wait}.
The bias is $-100$, the most negative value vLLM accepts, applied to the whole generation, with a 32{,}768-token budget as in the paper (24{,}576 on GPQA-Diamond) and the sampling settings of Appendix~\ref{app:exp}.
For Qwen3-32B over five seeds, the paper reports chains 13--16\% shorter on AMC 2023, AIME 2024 and AIME 2025 with accuracy changes of $-0.8$, $+2.0$ and $-2.7$ points, and on GPQA-Diamond chains 15\% shorter at $-6.1$ points.
Our plain-decoding chains have the same lengths as the paper's on the three mathematics benchmarks (within 5\%). NoWait removes more in our runs (21--24\%), and its accuracy cost falls on AIME 2025 and HMMT, while GPQA-Diamond loses 0.3 points (Table~\ref{tab:full}).
An independent run of NoWait on Qwen3-32B in the ThinkBrake paper~\citep{song2025thinkbrake} loses 9 points on AIME 2024 and 12 on AIME 2025 at 11--12\% fewer tokens. Our runs show the same direction, with the largest loss on AIME 2025.
\paragraph{DEER.}
DEER only decides when the CoT ends. A DEER run is therefore the plain-decoding run of the same problem and seed, cut at the first point where the rule fires, followed by \closetag{} and an answer sampled with the budget that remains. A run in which the rule never fires is the plain run itself. We evaluate it on the stored plain-decoding runs of Table~\ref{tab:main}, in the same engine and with the same sampler and seeds.
For DEER we follow the authors' released vLLM implementation with its default settings. The transition point is the string \emph{Wait}. At each of the first ten, the text before it is extended with the inducer \texttt{\textbackslash n**Final Answer**\textbackslash n\textbackslash boxed} (on GPQA-Diamond, the released variant \texttt{\ldots I believe the final answer, rather than the option, is \textbackslash boxed}) and a trial answer of at most 20 tokens is decoded greedily and stopped at \closetag{}. Its confidence is the geometric mean of the token probabilities from the second trial token on, and is zero unless the trial reached \closetag{}. The block is closed with \texttt{\textbackslash n</think>\textbackslash n\textbackslash n} when the confidence exceeds $0.95$, and in any case once the thinking reaches $0.8\times$ the budget (the released \texttt{think\_ratio}), at the segment boundary that precedes it by at most 50 tokens. After the tenth trial no further trial is made, and the chain runs to its own end or to the $0.8\times$ cap. The released code covers mathematics and GPQA. On LiveCodeBench the inducer opens a \texttt{python} code block and the trial may run to 1{,}024 tokens, with the same \closetag{} requirement.
The DEER paper also reports Qwen3-32B under the same sampler (temperature $0.6$, top-$p$ $0.95$, budget 32{,}768; its Table~8): thinking tokens fall by 10\% on AIME 2024, 9\% on AIME 2025, 22\% on AMC 2023 and 11\% on OlympiadBench, with accuracy within one point of plain decoding. Our runs give 8\%, 10\%, 20\% and 12\%, with accuracy within 2.2 points (Table~\ref{tab:full}).
\paragraph{ThinkBrake.}
ThinkBrake~\citep{song2025thinkbrake} ends the CoT at a sentence boundary when the margin between the most likely next token other than \closetag{} and \closetag{}, on the temperature-scaled logits, is at most $\tau$. Its paper uses $\tau{=}0.1$ for every Qwen3 model.
The paper does not say in which state the margin is read. Its only released code (for its tool-use benchmark) appends a period and a single newline to each prefix and reads \closetag{} there, which is the state in which Qwen3 closes its CoT on its own. In our runs every natural close follows a single-newline token, whereas a sentence inside the CoT mostly ends in a blank-line token, after which \closetag{} is never near the top. Read in the state the model actually wrote, the rule therefore does not fire.
We therefore follow the released code. At every sentence boundary, a token that ends a sentence and is followed by whitespace, or a token that ends with a newline, the sentence's last token is replaced by its single-newline form (\texttt{.\textbackslash n\textbackslash n} becomes \texttt{.\textbackslash n}) and the margin is read there. Where the rule fires, the run keeps its CoT up to that sentence, followed by the single newline, \closetag{} and a freshly sampled answer. Like DEER, ThinkBrake is evaluated on the stored plain-decoding runs.
It removes 21\% of the thinking tokens over the seven benchmarks at a cost of 2.3 points: 37\% on GPQA-Diamond and 44\% on LiveCodeBench, where it costs 2.0 and 4.7 points, and 3--4\% on the five others (Table~\ref{tab:full}). The paper reports 4--6\% fewer thinking tokens on AIME 2024 and 2025 and 20\% on GPQA-Diamond for Qwen3-32B with a 16k budget.
\paragraph{ASC.}
The published version of ASC learns the steering vector by gradient descent, which requires an additional training run. We follow the training-free version in the authors' released code, which adds the mean of the last-token differences between verbose and concise CoTs, scaled by a strength, to the output of one decoder layer at every forward pass, so at every generated token and at the last prompt token. The calibration set is the first 100 problems of MATH-500 in an order that cycles through its subjects. For each, Qwen3-32B writes one CoT under its plain instruction and one under the brief thinker's sentence, with the sampling settings of Appendix~\ref{app:exp} and a 16{,}384-token budget, and both are read back after the plain instruction. The authors write the concise CoTs with GPT-4o, and we use the large model's own. The vector is added after layer 57. Its strength $\rho$ is the norm of the added vector relative to the mean norm of the hidden state at that layer, and $\rho{=}0.2$ is chosen on AIME 2026 (Table~\ref{tab:alpha26}).

\FloatBarrier
\section{Experimental setup details}\label{app:exp}
\paragraph{Benchmarks, budgets and licenses.}
\begin{table}[H]
\centering\small
\caption{Benchmarks: problems $\times$ seeds per problem, thinking budget in tokens, answer type and license.}
\label{tab:bench}
\setlength{\tabcolsep}{3pt}\footnotesize
\resizebox{\columnwidth}{!}{%
\begin{tabular}{lrrll}
\toprule
Benchmark & $n\times$seeds & Budget & Answer & License \\
\midrule
AIME 2024 & $30\times3$ & 32k & integer & MAA \\
AIME 2025 & $30\times3$ & 32k & integer & MAA \\
HMMT Feb.\ 2025 & $30\times3$ & 32k & expression & HMMT \\
AMC 2023 & $40\times3$ & 32k & integer & MAA \\
GPQA-Diamond & $198\times3$ & 24k & letter & CC BY 4.0 \\
OlympiadBench & $100\times3$ & 32k & expression & Apache 2.0 \\
LiveCodeBench & $100\times3$ & 32k & program & MIT \\
\bottomrule
\end{tabular}}
\end{table}
AIME 2024 and AIME 2025 are used in full (both contests of each year, 30 problems), and so are HMMT February 2025 (30), AMC 2023 (40) and GPQA-Diamond (198).
OlympiadBench's 100 problems and LiveCodeBench's 100 are drawn round-robin across subjects and across difficulty levels respectively.
The thinking budget is 32{,}768 tokens except 24{,}576 on GPQA-Diamond.
Under plain decoding the share of runs that reach it is 13\% on AIME 2025, 3\% on AIME 2024, 4\% on HMMT, 0\% on AMC 2023, 0.3\% on GPQA-Diamond, 4\% on OlympiadBench and 2\% on LiveCodeBench.
LiveCodeBench problems are those of release 6 (January to April 2025), GPQA-Diamond is the 198-question diamond split, and OlympiadBench's problems are its text-only mathematics problems in English, so every problem in this paper is in English.
The competition problems (AIME, AMC, HMMT) are the property of their organizers and are used for evaluation only, as in the work we compare with. GPQA-Diamond is released under CC BY 4.0, OlympiadBench under Apache 2.0 and LiveCodeBench under MIT.

\paragraph{Scoring.}
The visible answer after \closetag{} is scored, never the CoT.
Integer answers (AIME, AMC) must match exactly.
LaTeX answers (HMMT, OlympiadBench) are compared with \texttt{math-verify}~\citep{mathverify}. An answer whose parts are boxed one by one is compared, as one list, with an answer key of several parts; an answer key that names what it gives ($m_{\max}=n^2-n-1$) is compared by its right side; and functions are compared after renaming their variable, so that $f(n)=n$ matches $f(z)=z$. We correct one OlympiadBench answer key, which gives $(10^3+1)^2$ itself where the problem asks for the sum of its digits.
Multiple-choice answers are matched by letter, and an answer that spells out an option's text or value, also in \LaTeX{}, is mapped back to its letter first.
LiveCodeBench programs are executed against the test cases.
A run that reaches its budget without ending its CoT scores zero.

\paragraph{Sampling, seeds and intervals.}
All settings sample with temperature $0.6$, top-$p$ $0.95$ and top-$k$ $20$, the settings Qwen3 recommends for thinking mode.
The seed of a run is derived from the problem index, the seed index and the setting, and the engine seeds the sampler of each run with it.
Intervals are percentile bootstraps with 4{,}000 resamples over problems.

\paragraph{Compute.}
All runs use vLLM 0.29, with the small model's two passes beside the large model on the same GPUs.
The experiments in this paper took about 2{,}000 A100 80GB GPU hours and 300 B200 GPU hours.

\FloatBarrier
\section{Strength and small-model size}\label{app:strength}
\begin{table}[H]
\centering\footnotesize
\caption{AIME 2025, 30 problems, three seeds each: \method{} (0.6B) at three strengths and at a negative one ($\alpha{=}-5$ adds the verbose minus the brief thinker). Accuracy, share of runs reaching the budget, mean thinking tokens, change in thinking tokens against plain decoding over all problems with its 95\% interval, and accuracy change in points.}
\label{tab:strength}
\setlength{\tabcolsep}{3pt}
\resizebox{\columnwidth}{!}{%
\begin{tabular}{l rrr l r}
\toprule
Setting & Acc. & Budget & Tokens & Token change & Acc.\ change \\
\midrule
$\alpha{=}-5$ & 73.3 & 21.1 & 19,676 & $+26\%$ \ci{+17}{+39} & $+1.1$ \\
Plain decoding & 72.2 & 13.3 & 15,606 & -- & -- \\
$\alpha{=}2.5$ & 74.4 & 4.4 & 11,964 & $-23\%$ \ci{-29}{-17} & $+2.2$ \\
$\alpha{=}5$ (default) & 72.2 & 1.1 & 11,092 & $-29\%$ \ci{-37}{-21} & $0.0$ \\
$\alpha{=}10$ & 64.4 & 3.3 & 9,999 & $-36\%$ \ci{-44}{-28} & $-7.8$ \\
\bottomrule
\end{tabular}}
\end{table}

\begin{table}[H]
\centering\footnotesize
\caption{AIME 2025, three seeds: \method{} with three small models of the same family at $\alpha{=}5$, and with Qwen3-32B itself as both thinkers at $\alpha{=}1$. Accuracy, share of runs reaching the budget, mean thinking tokens, change in thinking tokens over all problems, and accuracy change in points.}
\label{tab:size}
\setlength{\tabcolsep}{3pt}
\resizebox{\columnwidth}{!}{%
\begin{tabular}{l rrr l r}
\toprule
Small model & Acc. & Budget & Tokens & Token change & Acc.\ change \\
\midrule
Qwen3-0.6B & 72.2 & 1.1 & 11,092 & $-29\%$ \ci{-37}{-21} & $0.0$ \\
Qwen3-1.7B & 50.0 & 0.0 & 6,584 & $-58\%$ \ci{-64}{-51} & $-22.2$ \\
Qwen3-4B & 52.2 & 0.0 & 6,461 & $-59\%$ \ci{-64}{-53} & $-20.0$ \\
\midrule
Qwen3-32B itself & 73.3 & 4.4 & 11,590 & $-26\%$ \ci{-32}{-20} & $+1.1$ \\
\bottomrule
\end{tabular}}
\end{table}

\paragraph{Choosing the strength.}\label{app:alpha26}
AIME 2026 was released after Qwen3 and appears nowhere else in this paper. Table~\ref{tab:alpha26} sweeps $\alpha$ on its 30 problems with three seeds each: \method{} removes 21--23\% of the thinking tokens at $\alpha\le3.5$, 29\% at $\alpha{=}5$ and 32--33\% above it, while accuracy stays within noise of plain decoding up to $\alpha{=}7$ and falls by 4.4 points at $\alpha{=}10$. $\alpha{=}5$ is where the removal levels off. For ASC we sweep the strength the same way, and $\rho{=}0.2$ removes the most, 9\%, while larger strengths remove less (Table~\ref{tab:alpha26}).
\begin{table}[H]
\centering\footnotesize
\caption{AIME 2026 (30 problems, released after Qwen3), three seeds each: plain decoding, \method{} (0.6B) at five strengths and ASC at three. Accuracy, share of runs reaching the budget, mean thinking tokens, change in thinking tokens against plain decoding over all problems with its 95\% interval, and accuracy change in points.}
\label{tab:alpha26}
\setlength{\tabcolsep}{3pt}
\resizebox{\columnwidth}{!}{%
\begin{tabular}{l rrr l r}
\toprule
Setting & Acc. & Budget & Tokens & Token change & Acc.\ change \\
\midrule
Plain decoding & 68.9 & 8.9 & 13,543 & -- & -- \\
$\alpha{=}2.5$ & 76.7 & 3.3 & 10,474 & $-23\%$ \ci{-30}{-16} & $+7.8$ \\
$\alpha{=}3.5$ & 67.8 & 5.6 & 10,662 & $-21\%$ \ci{-30}{-13} & $-1.1$ \\
$\alpha{=}5$ (default) & 72.2 & 4.4 & 9,605 & $-29\%$ \ci{-38}{-20} & $+3.3$ \\
$\alpha{=}7$ & 68.9 & 3.3 & 9,275 & $-32\%$ \ci{-40}{-22} & $0.0$ \\
$\alpha{=}10$ & 64.4 & 2.2 & 9,135 & $-33\%$ \ci{-42}{-23} & $-4.4$ \\
\midrule
ASC $\rho{=}0.2$ (default) & 78.9 & 5.6 & 12,267 & $-9\%$ \ci{-16}{-2} & $+10.0$ \\
ASC $\rho{=}0.4$ & 74.4 & 4.4 & 12,446 & $-8\%$ \ci{-18}{+1} & $+5.6$ \\
ASC $\rho{=}0.8$ & 67.8 & 4.4 & 13,478 & $0\%$ \ci{-8}{+9} & $-1.1$ \\
\bottomrule
\end{tabular}}
\end{table}

\FloatBarrier
\section{Every setting on every benchmark}\label{app:full}
\begin{figure*}[htbp]
\centering
\includegraphics[width=\textwidth]{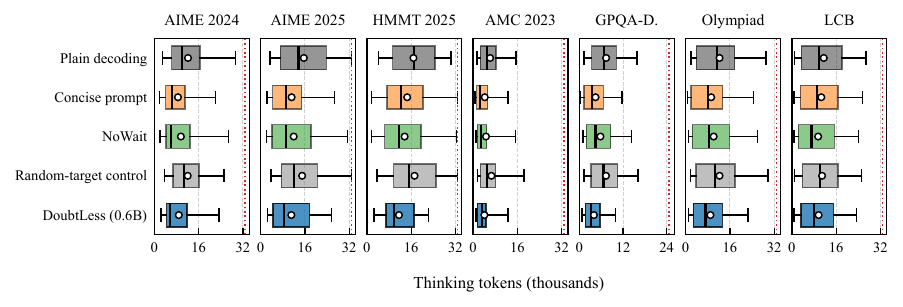}
\caption{Thinking length under each method on each benchmark: box, quartiles; whiskers, 5th to 95th percentile; dot, mean; dotted line, the budget.}
\label{fig:boxall}
\end{figure*}

For each problem $i$, let $\ell_i$ and $\ell'_i$ be the mean thinking length over its runs under plain decoding and under a method. The change of the method against plain decoding is $\bigl(\sum_i \ell'_i-\sum_i \ell_i\bigr)/\sum_i \ell_i$, with a percentile bootstrap over problems for the interval. It uses every problem, so it does not depend on which problems a method solves.
Table~\ref{tab:full} lists, for the five settings, DEER, ThinkBrake and ASC on the seven benchmarks, the accuracy, the share of runs that reach the budget, the mean thinking length, this change, and the change over the problems both the method and plain decoding solved in every run, the measure used in earlier work.
Figure~\ref{fig:boxall} draws the thinking length under each of the five settings on every benchmark.
\begin{table*}[p]
\centering\footnotesize
\caption{Every setting on every benchmark: accuracy and its change against plain decoding in points (paired over problems), share of runs that reach the budget (for DEER and ThinkBrake, in the answer after the thinking is closed), mean thinking tokens, change in thinking tokens against plain decoding over all problems, and the same change over the problems that both the setting and plain decoding solved in every run (their number in parentheses). 95\% bootstrap intervals over problems. The last block pools the seven benchmarks.}
\label{tab:full}
\setlength{\tabcolsep}{3pt}\renewcommand{\arraystretch}{0.8}
\scalebox{0.92}{%
\begin{tabular}{l r l rr l l}
\toprule
Setting & Acc.\ (\%) & Acc.\ change & Budget (\%) & Tokens & Token change, all & Token change, both solved \\
\midrule
\multicolumn{7}{l}{\emph{AIME 2024} (30 problems, 3 seeds each, budget 32,768)}\\
Plain decoding & 81.1 & -- & 3.3 & 12,152 & -- & -- \\
Concise prompt & 83.3 & $+2.2$ \ci{-5.6}{+8.9} & 1.1 & 8,517 & $-30\%$ \ci{-38}{-21} & $-28\%$ \ci{-37}{-20} (20) \\
NoWait & 77.8 & $-3.3$ \ci{-12.2}{+4.4} & 2.2 & 9,575 & $-21\%$ \ci{-32}{-10} & $-26\%$ \ci{-40}{-13} (18) \\
\method{} (0.6B) & 83.3 & $+2.2$ \ci{-6.7}{+11.1} & 0.0 & 8,816 & $-27\%$ \ci{-37}{-18} & $-34\%$ \ci{-41}{-27} (18) \\
Random-target control & 81.1 & $0.0$ \ci{-7.8}{+7.8} & 1.1 & 12,036 & $-1\%$ \ci{-10}{+9} & $+8\%$ \ci{-1}{+18} (18) \\
DEER & 82.2 & $+1.1$ \ci{+0.0}{+3.3} & 0.0 & 11,133 & $-8\%$ \ci{-12}{-5} & $-11\%$ \ci{-19}{-5} (20) \\
ThinkBrake & 80.0 & $-1.1$ \ci{-3.3}{+0.0} & 2.2 & 11,635 & $-4\%$ \ci{-9}{-1} & $-2\%$ \ci{-3}{-1} (20) \\
ASC & 82.2 & $+1.1$ \ci{-5.6}{+7.8} & 0.0 & 10,734 & $-12\%$ \ci{-20}{-3} & $-6\%$ \ci{-16}{+7} (18) \\
\addlinespace[1pt]
\multicolumn{7}{l}{\emph{AIME 2025} (30 problems, 3 seeds each, budget 32,768)}\\
Plain decoding & 72.2 & -- & 13.3 & 15,606 & -- & -- \\
Concise prompt & 75.6 & $+3.3$ \ci{-5.6}{+12.2} & 1.1 & 11,180 & $-28\%$ \ci{-37}{-19} & $-23\%$ \ci{-39}{-5} (16) \\
NoWait & 57.8 & $-14.4$ \ci{-24.4}{-4.4} & 5.6 & 11,939 & $-23\%$ \ci{-31}{-15} & $-38\%$ \ci{-48}{-24} (9) \\
\method{} (0.6B) & 72.2 & $0.0$ \ci{-11.1}{+10.0} & 1.1 & 11,092 & $-29\%$ \ci{-37}{-21} & $-32\%$ \ci{-39}{-26} (15) \\
Random-target control & 75.6 & $+3.3$ \ci{-3.3}{+11.1} & 14.4 & 14,992 & $-4\%$ \ci{-12}{+3} & $-2\%$ \ci{-8}{+6} (16) \\
DEER & 74.4 & $+2.2$ \ci{+0.0}{+5.6} & 4.4 & 14,087 & $-10\%$ \ci{-14}{-6} & $-8\%$ \ci{-17}{-3} (18) \\
ThinkBrake & 72.2 & $0.0$ \ci{-3.3}{+3.3} & 10.0 & 15,091 & $-3\%$ \ci{-6}{-1} & $-1\%$ \ci{-2}{+0} (17) \\
ASC & 73.3 & $+1.1$ \ci{-7.8}{+10.0} & 11.1 & 14,341 & $-8\%$ \ci{-15}{+1} & $-15\%$ \ci{-21}{-7} (15) \\
\addlinespace[1pt]
\multicolumn{7}{l}{\emph{HMMT Feb.\ 2025} (30 problems, 3 seeds each, budget 32,768)}\\
Plain decoding & 53.3 & -- & 4.4 & 16,817 & -- & -- \\
Concise prompt & 46.7 & $-6.7$ \ci{-16.7}{+3.3} & 10.0 & 14,551 & $-13\%$ \ci{-23}{-4} & $-34\%$ \ci{-55}{-4} (6) \\
NoWait & 46.7 & $-6.7$ \ci{-14.4}{+1.1} & 6.7 & 13,630 & $-19\%$ \ci{-25}{-13} & $-34\%$ \ci{-49}{-23} (6) \\
\method{} (0.6B) & 56.7 & $+3.3$ \ci{-3.3}{+11.1} & 0.0 & 11,586 & $-31\%$ \ci{-38}{-25} & $-29\%$ \ci{-49}{-11} (9) \\
Random-target control & 53.3 & $0.0$ \ci{-10.0}{+8.9} & 10.0 & 17,201 & $+2\%$ \ci{-5}{+10} & $+9\%$ \ci{-4}{+27} (6) \\
DEER & 53.3 & $0.0$ \ci{+0.0}{+0.0} & 2.2 & 15,739 & $-6\%$ \ci{-9}{-4} & $-6\%$ \ci{-18}{-1} (9) \\
ThinkBrake & 52.2 & $-1.1$ \ci{-3.3}{+0.0} & 3.3 & 16,144 & $-4\%$ \ci{-6}{-2} & $-1\%$ \ci{-3}{+0} (9) \\
ASC & 51.1 & $-2.2$ \ci{-13.3}{+7.8} & 11.1 & 17,303 & $+3\%$ \ci{-7}{+15} & $-10\%$ \ci{-29}{+11} (8) \\
\addlinespace[1pt]
\multicolumn{7}{l}{\emph{AMC 2023} (40 problems, 3 seeds each, budget 32,768)}\\
Plain decoding & 96.7 & -- & 0.0 & 6,130 & -- & -- \\
Concise prompt & 95.8 & $-0.8$ \ci{-6.7}{+5.0} & 0.8 & 4,203 & $-31\%$ \ci{-42}{-20} & $-39\%$ \ci{-46}{-31} (32) \\
NoWait & 94.2 & $-2.5$ \ci{-5.8}{+0.8} & 0.8 & 4,651 & $-24\%$ \ci{-39}{-7} & $-41\%$ \ci{-46}{-34} (35) \\
\method{} (0.6B) & 97.5 & $+0.8$ \ci{-1.7}{+3.3} & 0.0 & 4,063 & $-34\%$ \ci{-41}{-27} & $-33\%$ \ci{-41}{-25} (36) \\
Random-target control & 98.3 & $+1.7$ \ci{-1.7}{+5.0} & 0.8 & 6,605 & $+8\%$ \ci{-1}{+17} & $+3\%$ \ci{-3}{+9} (36) \\
DEER & 97.5 & $+0.8$ \ci{+0.0}{+2.5} & 0.0 & 4,875 & $-20\%$ \ci{-29}{-13} & $-20\%$ \ci{-29}{-12} (37) \\
ThinkBrake & 97.5 & $+0.8$ \ci{+0.0}{+2.5} & 0.0 & 5,919 & $-3\%$ \ci{-5}{-2} & $-4\%$ \ci{-5}{-2} (37) \\
ASC & 98.3 & $+1.7$ \ci{+0.0}{+4.2} & 0.0 & 5,458 & $-11\%$ \ci{-19}{-3} & $-15\%$ \ci{-21}{-8} (37) \\
\addlinespace[1pt]
\multicolumn{7}{l}{\emph{GPQA-Diamond} (198 problems, 3 seeds each, budget 24,576)}\\
Plain decoding & 68.7 & -- & 0.3 & 7,354 & -- & -- \\
Concise prompt & 67.7 & $-1.0$ \ci{-4.4}{+2.4} & 0.0 & 4,429 & $-40\%$ \ci{-43}{-37} & $-48\%$ \ci{-53}{-43} (97) \\
NoWait & 68.4 & $-0.3$ \ci{-3.5}{+3.0} & 1.0 & 5,782 & $-21\%$ \ci{-25}{-17} & $-28\%$ \ci{-33}{-23} (92) \\
\method{} (0.6B) & 66.3 & $-2.4$ \ci{-5.6}{+0.8} & 0.0 & 3,995 & $-46\%$ \ci{-48}{-43} & $-43\%$ \ci{-48}{-39} (97) \\
Random-target control & 70.0 & $+1.3$ \ci{-1.7}{+4.4} & 0.5 & 7,319 & $0\%$ \ci{-3}{+2} & $-1\%$ \ci{-6}{+3} (99) \\
DEER & 68.7 & $0.0$ \ci{-0.5}{+0.5} & 0.0 & 7,200 & $-2\%$ \ci{-3}{-1} & $-4\%$ \ci{-6}{-2} (110) \\
ThinkBrake & 66.7 & $-2.0$ \ci{-5.2}{+1.3} & 0.7 & 4,650 & $-37\%$ \ci{-41}{-33} & $-23\%$ \ci{-27}{-19} (95) \\
ASC & 69.9 & $+1.2$ \ci{-1.7}{+4.0} & 1.0 & 7,058 & $-4\%$ \ci{-7}{-1} & $-6\%$ \ci{-11}{-1} (103) \\
\addlinespace[1pt]
\multicolumn{7}{l}{\emph{OlympiadBench} (100 problems, 3 seeds each, budget 32,768)}\\
Plain decoding & 70.7 & -- & 3.7 & 12,246 & -- & -- \\
Concise prompt & 68.7 & $-2.0$ \ci{-6.0}{+2.0} & 1.7 & 9,227 & $-25\%$ \ci{-29}{-20} & $-36\%$ \ci{-40}{-31} (56) \\
NoWait & 69.7 & $-1.0$ \ci{-5.0}{+2.7} & 1.7 & 10,117 & $-17\%$ \ci{-23}{-11} & $-16\%$ \ci{-26}{-7} (54) \\
\method{} (0.6B) & 69.0 & $-1.7$ \ci{-6.3}{+2.7} & 1.0 & 8,951 & $-27\%$ \ci{-31}{-22} & $-32\%$ \ci{-38}{-25} (55) \\
Random-target control & 70.7 & $0.0$ \ci{-3.7}{+3.3} & 2.0 & 12,239 & $0\%$ \ci{-5}{+5} & $-3\%$ \ci{-8}{+1} (57) \\
DEER & 71.3 & $+0.7$ \ci{-0.7}{+2.0} & 0.7 & 10,762 & $-12\%$ \ci{-17}{-8} & $-15\%$ \ci{-23}{-9} (63) \\
ThinkBrake & 67.3 & $-3.3$ \ci{-7.0}{-0.3} & 3.0 & 11,770 & $-4\%$ \ci{-5}{-3} & $-3\%$ \ci{-5}{-2} (60) \\
ASC & 69.0 & $-1.7$ \ci{-6.3}{+2.3} & 2.3 & 11,708 & $-4\%$ \ci{-9}{+1} & $-5\%$ \ci{-11}{+2} (55) \\
\addlinespace[1pt]
\multicolumn{7}{l}{\emph{LiveCodeBench} (100 problems, 3 seeds each, budget 32,768)}\\
Plain decoding & 66.3 & -- & 2.0 & 11,511 & -- & -- \\
Concise prompt & 65.7 & $-0.7$ \ci{-4.3}{+2.7} & 1.3 & 10,628 & $-8\%$ \ci{-12}{-3} & $-9\%$ \ci{-14}{-5} (52) \\
NoWait & 62.3 & $-4.0$ \ci{-8.0}{-0.3} & 0.3 & 9,489 & $-18\%$ \ci{-22}{-13} & $-25\%$ \ci{-30}{-21} (48) \\
\method{} (0.6B) & 66.0 & $-0.3$ \ci{-4.0}{+3.3} & 1.0 & 9,659 & $-16\%$ \ci{-21}{-11} & $-15\%$ \ci{-22}{-7} (51) \\
Random-target control & 66.3 & $0.0$ \ci{-4.0}{+3.7} & 1.0 & 10,901 & $-5\%$ \ci{-9}{-1} & $0\%$ \ci{-5}{+5} (52) \\
DEER & 61.7 & $-4.7$ \ci{-7.7}{-2.0} & 0.7 & 9,509 & $-17\%$ \ci{-24}{-12} & $-10\%$ \ci{-16}{-5} (46) \\
ThinkBrake & 61.7 & $-4.7$ \ci{-8.3}{-1.3} & 0.7 & 6,408 & $-44\%$ \ci{-49}{-39} & $-21\%$ \ci{-25}{-18} (49) \\
ASC & 66.0 & $-0.3$ \ci{-4.7}{+4.0} & 1.3 & 11,080 & $-4\%$ \ci{-9}{+2} & $-4\%$ \ci{-11}{+3} (51) \\
\addlinespace[1pt]
\multicolumn{7}{l}{\emph{All seven benchmarks, pooled} (528 problems, three seeds each)}\\
Plain decoding & 70.8 & -- & 2.4 & 10,254 & -- & -- \\
Concise prompt & 69.8 & $-1.0$ \ci{-2.9}{+0.9} & 1.3 & 7,686 & $-25\%$ \ci{-27}{-23} & $-33\%$ \ci{-37}{-30} (279) \\
NoWait & 68.1 & $-2.7$ \ci{-4.5}{-0.8} & 1.6 & 8,231 & $-20\%$ \ci{-22}{-17} & $-27\%$ \ci{-30}{-23} (262) \\
\method{} (0.6B) & 69.9 & $-0.9$ \ci{-2.8}{+1.0} & 0.4 & 7,120 & $-31\%$ \ci{-33}{-28} & $-33\%$ \ci{-36}{-30} (281) \\
Random-target control & 71.6 & $+0.8$ \ci{-0.9}{+2.5} & 2.3 & 10,140 & $-1\%$ \ci{-3}{+1} & $0\%$ \ci{-3}{+2} (284) \\
DEER & 70.3 & $-0.5$ \ci{-1.2}{+0.2} & 0.6 & 9,236 & $-10\%$ \ci{-12}{-8} & $-10\%$ \ci{-13}{-8} (303) \\
ThinkBrake & 68.4 & $-2.3$ \ci{-3.9}{-0.8} & 1.8 & 8,071 & $-21\%$ \ci{-24}{-19} & $-10\%$ \ci{-12}{-8} (287) \\
ASC & 71.0 & $+0.2$ \ci{-1.6}{+2.0} & 2.3 & 9,784 & $-5\%$ \ci{-7}{-2} & $-7\%$ \ci{-10}{-5} (287) \\
\bottomrule
\end{tabular}}
\end{table*}

\section{Other model families}\label{app:families}
\begin{table}[t]
\centering\footnotesize
\caption{\method{} removes about three times as much thinking as the concise prompt on EXAONE-4.0 and Granite-4.2, with unchanged accuracy. Same instructions and strength for every family. AIME 2025 and AMC 2023, 70 problems, three seeds each; each large model is steered by the small model below it. Accuracy (\%) and thinking-token change against plain decoding, the latter with 95\% intervals.}
\label{tab:families}
\setlength{\tabcolsep}{3pt}
\resizebox{\columnwidth}{!}{%
\begin{tabular}{@{}l r rr rr@{}}
\toprule
 & Plain & \multicolumn{2}{c}{Concise} & \multicolumn{2}{c}{\method{}} \\
\cmidrule(lr){2-2}\cmidrule(lr){3-4}\cmidrule(l){5-6}
Model & Acc. & Acc. & Tokens & Acc. & Tokens \\
\midrule
Qwen3-32B & 86.2 & 87.1 & $-29\%$ & 86.7 & $-31\%$ \\
\hspace{0.6em}{\scriptsize\emph{Qwen3-0.6B}} & & & \ci{-36}{-22} & & \ci{-36}{-25} \\
\addlinespace[2pt]
EXAONE-4.0-32B & 88.6 & 88.6 & $-8\%$ & 90.5 & $-25\%$ \\
\hspace{0.6em}{\scriptsize\emph{EXAONE-4.0-1.2B}} & & & \ci{-11}{-4} & & \ci{-31}{-19} \\
\addlinespace[2pt]
Granite-4.2-30B & 89.0 & 88.1 & $-14\%$ & 91.4 & $-40\%$ \\
\hspace{0.6em}{\scriptsize\emph{Granite-4.2-3B}} & & & \ci{-21}{-9} & & \ci{-48}{-32} \\
\bottomrule
\end{tabular}}
\end{table}

EXAONE-4.0-32B and EXAONE-4.0-1.2B share the EXAONE tokenizer and open the CoT with the same tag as Qwen3.
Granite-4.2-30B and Granite-4.2-3B~\citep{ibm2026granite42} share the Granite tokenizer.
The instruction and the two thinker sentences are identical to Appendix~\ref{app:prompts} for all three families.
The two other families use the sampling settings, thinking budget and strength ($\alpha{=}5$) of the main table, with three seeds per problem, and each family's own chat template with thinking enabled.
The concise prompt in Table~\ref{tab:families} is the setting of the main experiments, the brief thinker's sentence appended to the large model's own instruction, with three seeds per problem.

\section{Where the removed tokens come from}\label{app:where}
Each chain is split at the first automatic match to the correct answer in the CoT: for integer answers the first occurrence of the correct value as a standalone number, and for multiple-choice benchmarks the first occurrence of the correct letter in a phrase such as ``the answer is B''. Problems whose answer is a LaTeX expression are not split.
Table~\ref{tab:where} gives, for AIME 2025, the mean number of thinking tokens before and after that point under each setting, over the correct runs in which the answer can be located, and Table~\ref{tab:phrases} the frequency of some words per thousand thinking tokens. Figure~\ref{fig:removed} draws both.
Table~\ref{tab:whereall} repeats the split on every benchmark for the runs that both plain decoding and \method{} solve, so that the two methods are compared on the same runs. LiveCodeBench is absent because the model writes its program after the CoT, and HMMT and OlympiadBench cover their integer-answer problems only.
\begin{table}[t]
\centering\footnotesize
\caption{AIME 2025: mean thinking tokens before and after the first occurrence of the correct answer, over the correct runs in which the answer can be located (last column).}
\label{tab:where}
\setlength{\tabcolsep}{3.5pt}
\resizebox{\columnwidth}{!}{%
\begin{tabular}{lrrrr}
\toprule
Setting & Before & After & Share after & Runs \\
\midrule
Plain decoding & 7,624 & 3,553 & 32\% & 65 \\
Concise prompt & 6,096 & 2,551 & 30\% & 68 \\
NoWait & 5,478 & 1,334 & 20\% & 52 \\
\method{} (0.6B) & 6,023 & 1,710 & 22\% & 65 \\
Random-target control & 6,716 & 3,851 & 36\% & 68 \\
\bottomrule
\end{tabular}}
\end{table}

\begin{table}[t]
\centering\footnotesize
\caption{AIME 2025: occurrences per thousand thinking tokens of some words, pooled over all runs of a setting.}
\label{tab:phrases}
\setlength{\tabcolsep}{3pt}
\resizebox{\columnwidth}{!}{%
\begin{tabular}{lrrrrr}
\toprule
Word & Plain & Concise & NoWait & \method{} & Random \\
\midrule
Wait & 3.2 & 3.2 & 0.0 & 3.6 & 3.2 \\
Hmm & 0.1 & 0.1 & 0.0 & 0.1 & 0.1 \\
But & 2.7 & 2.7 & 0.0 & 3.5 & 2.8 \\
Alternatively & 1.1 & 1.2 & 0.0 & 1.3 & 1.0 \\
However & 1.5 & 1.4 & 0.0 & 1.3 & 1.6 \\
verify & 0.2 & 0.2 & 0.0 & 0.2 & 0.2 \\
double-check & 0.0 & 0.0 & 0.1 & 0.0 & 0.0 \\
Therefore & 6.1 & 5.2 & 6.8 & 5.3 & 6.2 \\
So & 2.5 & 3.1 & 3.2 & 3.2 & 2.8 \\
Let me & 4.0 & 4.3 & 6.3 & 3.8 & 3.9 \\
\bottomrule
\end{tabular}}
\end{table}

\begin{table*}[t]
\centering\small
\caption{Plain decoding and \method{} (0.6B) on the same runs, those that both settings solve and in which the answer can be located (integer answers as a standalone number, letters as the answer of a phrase). Mean thinking tokens before and after the first occurrence of the correct answer under each setting, the change of each part, and the share of the removed tokens that lie after the answer. HMMT and OlympiadBench cover their integer-answer problems only; on LiveCodeBench the program is written after the CoT, so no split is possible.}
\label{tab:whereall}
\setlength{\tabcolsep}{4pt}
\begin{tabular}{l r rr rr rr r}
\toprule
 & & \multicolumn{2}{c}{Plain decoding} & \multicolumn{2}{c}{\method{} (0.6B)} & \multicolumn{2}{c}{Change} & Removed \\
\cmidrule(lr){3-4}\cmidrule(lr){5-6}\cmidrule(lr){7-8}
Benchmark & Runs & Before & After & Before & After & Before & After & after answer \\
\midrule
AIME 2024 & 68 & 5,751 & 3,301 & 5,186 & 1,474 & $-10\%$ & $-55\%$ & 76\% \\
AIME 2025 & 56 & 6,310 & 3,700 & 5,183 & 1,630 & $-18\%$ & $-56\%$ & 65\% \\
HMMT Feb.\ 2025 & 19 & 4,000 & 4,973 & 3,573 & 1,382 & $-11\%$ & $-72\%$ & 89\% \\
AMC 2023 & 115 & 1,880 & 3,974 & 1,666 & 2,330 & $-11\%$ & $-41\%$ & 89\% \\
GPQA-Diamond & 346 & 2,833 & 2,799 & 1,991 & 1,179 & $-30\%$ & $-58\%$ & 66\% \\
OlympiadBench & 21 & 1,525 & 8,639 & 1,014 & 7,029 & $-34\%$ & $-19\%$ & 76\% \\
\bottomrule
\end{tabular}
\end{table*}

\paragraph{Where the two instructions disagree.}
For Figure~\ref{fig:disagree} the small model reads each correct plain-decoding chain on AIME 2025 twice, once after the brief thinker's prompt and once after the verbose thinker's, the two prompts \method{} uses.
At every position of the CoT we record the KL divergence from the brief to the verbose next-token distribution, computed over the brief thinker's 64 most likely tokens, and the difference between the two log-probabilities of the token the large model wrote next.
Positions are aligned at the first automatic match to the correct answer and averaged over runs in bins of 250 tokens, keeping the bins that at least ten runs reach. Panel (b) pools every token whose text, without surrounding spaces, is the given word.
The 39 correct runs of at most 12{,}000 thinking tokens enter, from 16 problems.
The shading of Figure~\ref{fig:direction} uses the same two readings: a token is shaded red when Equation~\eqref{eq:poe} down-weights it by a factor of five or more and green when it up-weights it by five or more, that is, when $|\log p_{\mathrm{brief}}-\log p_{\mathrm{verbose}}|\geq 0.2$ at $\alpha{=}5$ and temperature $0.6$.

\paragraph{Human validation of answer boundaries.}
Two annotators independently marked the first statement of the correct target answer in all 39 traces, allowing tentative statements and equivalent expressions while excluding incidental numbers.
We repeated the analysis using each annotation set separately.
The post-/pre-answer mean divergence ratios are 7.1 and 7.4, compared with 7.4 under automatic matching, and the share of the total divergence after the boundary is 84\% under all three.
Under either annotation set, mean divergence is larger after the boundary in all 39 traces.

\paragraph{Retained error-correcting verification.}\label{app:usefulcheck}
On AIME 2025 problem 11, a \method{} run initially rejects the positive root of $34y^2-y-32=0$, treating its rounded approximation as an excluded interval endpoint.
An algebraic check then uses $x=34y^2=y+32$ to establish that $32<x<33$, restoring the root.
The final answer includes both roots and uses their sum, $1/34$.
Both annotators verified the error, its correction, and its effect on the final sum.
This run uses 17{,}196 thinking tokens, compared with 20{,}866 under plain decoding.

\section{The self-debugging agent}\label{app:agent}
The agent solves each of the 164 HumanEval+ problems~\citep{liu2024evalplus} in at most two turns.
In the first turn the model receives the function to complete with the instruction ``Complete this Python function. Return the FULL function in one \texttt{python} block'', thinks, and writes the program. The program is run against the HumanEval+ tests.
If it fails, the model receives the test's error message with the instruction ``Your program failed the tests with this error: \ldots\ Fix the program. Return the FULL function in one \texttt{python} block'' and gets one more turn.
Every turn has its own CoT with a budget of 4{,}096 tokens and 1{,}024 tokens for the program. \method{} is applied inside each CoT exactly as in the single-turn runs, with the same two thinker sentences appended to the last user message, NoWait blocks the same 89 token ids, the concise prompt appends the brief sentence to the last user message, and the random-target control is applied as in the single-turn runs.
DEER and ThinkBrake run inside each turn as it is decoded, since an early exit in the first turn changes the program that the repair turn sees. DEER uses the code inducer and 1{,}024-token trials of LiveCodeBench (Appendix~\ref{app:impl}) and cuts the CoT at $0.8\times$ the turn's thinking budget. ThinkBrake first decodes the turn as plain decoding with the same seed, reads its rule at the sentence boundaries of that CoT as in the single-turn runs, and closes the CoT at the first boundary where it fires, followed by a freshly sampled program.
Under plain decoding 16--20\% of the turns reach the budget of 4{,}096 thinking tokens without closing their CoT and so write no program. \method{} halves the number of such turns, and DEER's cut leaves none.
The random-target control leaves the thinking per turn within 1\% of plain decoding and the pass rate within 1.2 points.
Three seeds per problem. The pass rate counts a problem as solved if the program passes all tests after the first or the second turn, averaged over seeds, and thinking tokens are averaged over turns and seeds.

\FloatBarrier
\section{Per-problem results on AIME 2025}\label{app:aime}
\begin{figure}[htbp]
\centering
\includegraphics[width=\columnwidth]{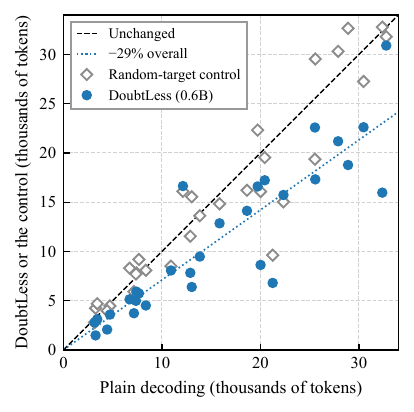}
\caption{\method{} removes a similar percentage on short and on long problems, and the random-target control stays on the diagonal. AIME 2025: mean thinking tokens of each problem over its three runs, under \method{} (0.6B) and the control against plain decoding. The dotted line is the overall change.}
\label{fig:perproblem}
\end{figure}

Figure~\ref{fig:perproblem} plots, for every problem, its mean thinking length under \method{} and under the random-target control against plain decoding.
The percentage removed does not depend on how long plain decoding thinks on a problem.
The problems lie along a line through the origin, the ones plain decoding finishes fastest and slowest lose a similar percentage (30\% and 29\% for the fastest and the slowest third), and the random-target control stays on the diagonal.
\method{} shifts the whole distribution of thinking length down, and the share of runs that reach the 32{,}768-token budget falls from 13\% to 1\% (Figure~\ref{fig:boxall}, Table~\ref{tab:full}).
Table~\ref{tab:aimeprob} lists every problem of AIME 2025 under the five settings: the mean thinking length over the three seeds and the number of seeds under which the answer was correct.
\begin{table*}[t]
\centering\small
\caption{AIME 2025, per problem: mean thinking tokens over the three seeds and (correct runs / runs) under each setting.}
\label{tab:aimeprob}
\setlength{\tabcolsep}{4pt}
\begin{tabular}{rr rrrrr}
\toprule
Problem & Answer & Plain decoding & Concise prompt & NoWait & \method{} (0.6B) & Random-target \\
\midrule
1 & 70 & 3,266 (3/3) & 1,532 (3/3) & 1,882 (3/3) & 1,479 (3/3) & 4,239 (3/3) \\
2 & 588 & 12,881 (3/3) & 10,077 (3/3) & 8,088 (3/3) & 7,827 (3/3) & 11,541 (3/3) \\
3 & 16 & 7,159 (3/3) & 4,369 (3/3) & 3,585 (3/3) & 3,726 (3/3) & 5,914 (3/3) \\
4 & 117 & 8,367 (3/3) & 5,931 (3/3) & 4,924 (2/3) & 4,514 (3/3) & 8,085 (3/3) \\
5 & 279 & 10,917 (3/3) & 6,712 (3/3) & 5,512 (2/3) & 8,070 (3/3) & 8,503 (3/3) \\
6 & 504 & 3,163 (3/3) & 2,495 (3/3) & 2,158 (3/3) & 2,791 (3/3) & 2,781 (3/3) \\
7 & 821 & 12,129 (1/3) & 9,160 (0/3) & 9,046 (1/3) & 16,634 (1/3) & 16,098 (2/3) \\
8 & 77 & 4,721 (3/3) & 2,695 (3/3) & 2,761 (2/3) & 3,589 (3/3) & 4,465 (3/3) \\
9 & 62 & 19,714 (3/3) & 8,776 (3/3) & 9,864 (3/3) & 16,599 (1/3) & 22,313 (2/3) \\
10 & 81 & 20,435 (3/3) & 19,167 (1/3) & 16,574 (1/3) & 17,223 (0/3) & 19,532 (2/3) \\
11 & 259 & 25,566 (2/3) & 19,240 (2/3) & 20,289 (1/3) & 17,315 (2/3) & 29,533 (1/3) \\
12 & 510 & 25,531 (1/3) & 11,788 (3/3) & 20,443 (1/3) & 22,579 (2/3) & 19,359 (2/3) \\
13 & 204 & 27,884 (0/3) & 14,204 (0/3) & 16,318 (0/3) & 21,178 (0/3) & 30,312 (0/3) \\
14 & 60 & 32,768 (0/3) & 25,255 (2/3) & 32,239 (0/3) & 30,898 (0/3) & 31,793 (0/3) \\
15 & 735 & 30,467 (0/3) & 16,114 (0/3) & 23,952 (0/3) & 22,607 (0/3) & 27,253 (0/3) \\
16 & 468 & 3,434 (3/3) & 3,284 (3/3) & 2,031 (3/3) & 3,117 (3/3) & 4,694 (3/3) \\
17 & 49 & 4,439 (3/3) & 2,002 (3/3) & 2,642 (3/3) & 2,070 (3/3) & 3,976 (3/3) \\
18 & 82 & 22,334 (1/3) & 17,321 (2/3) & 12,291 (2/3) & 15,731 (3/3) & 15,033 (3/3) \\
19 & 106 & 6,719 (3/3) & 4,006 (3/3) & 3,307 (3/3) & 5,137 (3/3) & 8,299 (3/3) \\
20 & 336 & 18,634 (3/3) & 17,015 (3/3) & 20,833 (2/3) & 14,123 (3/3) & 16,205 (3/3) \\
21 & 293 & 7,397 (3/3) & 4,807 (3/3) & 15,336 (2/3) & 5,922 (3/3) & 7,817 (3/3) \\
22 & 237 & 7,663 (2/3) & 6,409 (3/3) & 6,524 (3/3) & 5,738 (3/3) & 9,185 (3/3) \\
23 & 610 & 15,851 (3/3) & 14,868 (3/3) & 12,346 (2/3) & 12,853 (3/3) & 14,832 (3/3) \\
24 & 149 & 7,364 (3/3) & 3,748 (3/3) & 3,804 (1/3) & 4,989 (2/3) & 7,711 (3/3) \\
25 & 907 & 13,027 (3/3) & 11,294 (2/3) & 15,062 (1/3) & 6,391 (3/3) & 15,547 (3/3) \\
26 & 113 & 20,013 (2/3) & 12,529 (2/3) & 10,519 (0/3) & 8,631 (2/3) & 16,079 (3/3) \\
27 & 19 & 21,241 (2/3) & 6,542 (3/3) & 13,971 (2/3) & 6,804 (3/3) & 9,619 (2/3) \\
28 & 248 & 32,371 (0/3) & 27,215 (0/3) & 24,093 (0/3) & 15,969 (1/3) & 32,768 (0/3) \\
29 & 104 & 13,843 (3/3) & 21,053 (3/3) & 12,692 (3/3) & 9,476 (3/3) & 13,620 (3/3) \\
30 & 240 & 28,893 (0/3) & 25,781 (0/3) & 25,096 (0/3) & 18,772 (0/3) & 32,647 (0/3) \\
\bottomrule
\end{tabular}
\end{table*}

\FloatBarrier   %
\section{Worked examples}\label{app:example}
AIME 2025 problem 1 (I-1): \emph{Find the sum of all integer bases $b>9$ for which $17_b$ is a divisor of $97_b$.}
Both methods answer $70$.
Under plain decoding the CoT runs to \exPlainTok{} thinking tokens, and with \method{} (0.6B) it ends at \exOursTok.
Both are the runs of the first seed in the main experiment, decoded under the same sampling settings, and are shown in full.

\medskip\noindent\textbf{Plain decoding.}\par\nopagebreak
\begin{lstlisting}
Okay, so I need to find all integer bases $ b > 9 $ where the number $ 17_b $ divides $ 97_b $. Then, sum all those bases. Let me start by understanding what these numbers represent in base $ b $.

First, let me recall how numbers are represented in different bases. In base $ b $, the number $ 17_b $ means $ 1 \times b + 7 $, right? Similarly, $ 97_b $ would be $ 9 \times b + 7 $. So, translating both numbers to base 10 (decimal), we have:

$ 17_b = 1 \times b + 7 = b + 7 $

$ 97_b = 9 \times b + 7 = 9b + 7 $

So the problem is asking for bases $ b > 9 $ such that $ b + 7 $ divides $ 9b + 7 $. In other words, $ (9b + 7) $ divided by $ (b + 7) $ should result in an integer.

Let me write that as a division problem. Let me denote $ N = 9b + 7 $ and $ D = b + 7 $. We need $ D \mid N $, meaning there exists some integer $ k $ such that $ N = k \times D $. So:

$ 9b + 7 = k(b + 7) $

Our goal is to find all integers $ b > 9 $ such that this equation holds for some integer $ k $.

Alternatively, we can perform the division and find when the remainder is zero. Let me try polynomial division or algebraic manipulation.

Let me rearrange the equation:

$ 9b + 7 = k(b + 7) $

Let me solve for $ k $:

$ k = \frac{9b + 7}{b + 7} $

To find integer values of $ k $, we can perform the division. Let me do that.

Divide $ 9b + 7 $ by $ b + 7 $. Let me think about how to divide these linear expressions.

Dividing $ 9b + 7 $ by $ b + 7 $, the leading term is $ 9b / b = 9 $. Multiply divisor by 9: $ 9(b + 7) = 9b + 63 $. Subtract this from the dividend:

$ (9b + 7) - (9b + 63) = 9b + 7 - 9b - 63 = -56 $

Therefore, the division gives:

$ \frac{9b + 7}{b + 7} = 9 - \frac{56}{b + 7} $

So, for this to be an integer, $ \frac{56}{b + 7} $ must be an integer. Therefore, $ b + 7 $ must be a positive divisor of 56.

But since $ b > 9 $, then $ b + 7 > 9 + 7 = 16 $. Therefore, $ b + 7 $ is a divisor of 56 greater than 16.

So, first, let me list all positive divisors of 56, then pick those greater than 16, subtract 7 to get $ b $, check if $ b > 9 $, then sum them up.

First, factorize 56. 56 factors into 1, 2, 4, 7, 8, 14, 28, 56. Let me confirm:

Yes, 56 divided by 1 is 56

56 divided by 2 is 28

Divided by 4 is 14

Divided by 7 is 8

Then the divisors are 1, 2, 4, 7, 8, 14, 28, 56.

Now, we need divisors of 56 greater than 16. Looking at the list:

From the divisors: 28 and 56 are greater than 16.

Wait, 14 is 14, which is less than 16, so the next ones are 28 and 56. Therefore, $ b + 7 \in \{28, 56\} $. Therefore, solving for $ b $:

Case 1: $ b + 7 = 28 \implies b = 28 - 7 = 21 $

Case 2: $ b + 7 = 56 \implies b = 56 - 7 = 49 $

So the possible bases are 21 and 49. Both are greater than 9, so they satisfy the condition. Therefore, the sum is 21 + 49 = 70.

Wait, but let me check if there are any other divisors. Wait, 56's divisors are only up to 56. But perhaps negative divisors? But since $ b + 7 $ is a positive quantity (as base is positive integer greater than 9), we don't need to consider negative divisors.

But let me double-check my logic here. The key step was when I did the division and found that $ \frac{9b + 7}{b + 7} = 9 - \frac{56}{b + 7} $. Therefore, for this expression to be integer, $ \frac{56}{b + 7} $ must be integer. Therefore, $ b + 7 $ divides 56. Therefore, $ b + 7 $ is a positive divisor of 56, and since $ b > 9 $, $ b + 7 > 16 $. Therefore, divisors of 56 greater than 16 are 28 and 56, leading to b=21 and 49. Therefore, sum is 70.

But let me check if there are other divisors. For example, maybe 56 divided by something else? Wait, divisors are only those numbers. Let me check 56's divisors again:

Start from 1:

1 x 56 = 56

2 x 28 = 56

4 x 14 = 56

7 x 8 = 56

So divisors are 1, 2, 4, 7, 8, 14, 28, 56. So yes, only those. Therefore, divisors above 16 are 28 and 56. So that seems right.

But let me check if these bases actually work. Let me verify with b=21:

Compute $ 17_{21} = 1x21 +7 = 28 $

Compute $ 97_{21} = 9x21 +7 = 189 +7 = 196 $

Check if 28 divides 196: 196 / 28 = 7. Yes, that's an integer. So that works.

Now check b=49:

$ 17_{49} = 1x49 +7 = 56 $

$ 97_{49} = 9x49 +7 = 441 + 7 = 448 $

Check if 56 divides 448: 448 / 56 = 8. Yes, that's an integer. Perfect. So both bases work.

Therefore, the sum is 21 + 49 = 70.

Wait, but hold on. Let me think again. Is there a possibility that there are more divisors if we consider other factorizations?

Wait, for example, if I had considered that maybe $ b + 7 $ is a divisor of 56, but perhaps I missed some? But 56 is 2^3 * 7^1. Its divisors are indeed as listed. So no, there are no more divisors. So 28 and 56 are the only ones above 16.

But let me think again about the equation. Suppose I didn't do the division, but instead set up the congruence.

We need $ b + 7 $ divides $ 9b + 7 $. So, $ b + 7 \mid 9b + 7 $. Which implies that $ b + 7 \mid 9b + 7 - 9*(b + 7) $. Let me compute that:

$ 9b + 7 - 9*(b + 7) = 9b +7 -9b -63 = -56 $. Therefore, $ b + 7 \mid -56 $. Since divisibility is unaffected by sign, this implies $ b + 7 \mid 56 $. Which is the same conclusion as before. Therefore, same result. Therefore, same answer.

Therefore, the answer is 21 + 49 = 70. Therefore, the sum is 70.

But let me check if there's a possibility of other divisors if we consider that $ b + 7 $ could be less than or equal to 16 but somehow the original problem allows it? Wait, but the problem states that base $ b > 9 $. However, in the number $ 97_b $, the digits must be valid in base $ b $. In base $ b $, digits must be from 0 to $ b - 1 $. Therefore, in $ 97_b $, the digits are 9 and 7, so we must have $ b > 9 $, since digit 9 is present. Which is already given in the problem statement. Therefore, our initial condition $ b > 9 $ is correct.

But just to check, for example, if we had considered divisors between 7 and 16, but since $ b + 7 $ must be greater than 16, they are excluded. However, just for thoroughness, let me check if any of the smaller divisors would give a base $ b > 9 $. For example, take divisor 14: $ b + 7 = 14 \implies b = 7 $. But 7 is not greater than 9, so invalid. Similarly, divisor 8: $ b + 7 = 8 \implies b = 1 $, invalid. Similarly, 7: b=0, nope. 4: b=-3, nope. 2: b=-5, nope. 1: b=-6. So none of those work. The next divisor is 28, which gives b=21, which is valid. Then 56 gives 49. So yeah, only those two.

Therefore, the answer is 21 + 49 = 70. Therefore, the sum is 70. Therefore, I think that's the correct answer.

But let me try another approach to confirm. Let's suppose that $ b + 7 $ divides $ 9b + 7 $. Therefore, there exists integer k such that:

$ 9b + 7 = k(b + 7) $

Rearranging:

$ 9b + 7 = kb + 7k $

Bring all terms to left-hand side:

$ 9b + 7 - kb - 7k = 0 $

Factor:

$ b(9 - k) + 7(1 - k) = 0 $

Rearranged:

$ b(9 - k) = 7(k - 1) $

Therefore:

$ b = \frac{7(k - 1)}{9 - k} $

Since $ b > 9 $, and $ b $ must be a positive integer. Let me note that denominator is $ 9 - k $. Let me write this as:

Multiply numerator and denominator by -1:

$ b = \frac{7(1 - k)}{k - 9} $

Alternatively, maybe better to write:

Let me consider that $ 9 - k $ must divide $ 7(k - 1) $. Also, since $ b > 9 $, the right-hand side must be integer and greater than 9.

Alternatively, maybe express $ b $ as:

Let me write:

$ b = \frac{7(k - 1)}{9 - k} $

Let me note that $ 9 - k $ must divide 7(k - 1). Also, since b must be positive, the numerator and denominator must have same sign.

Case 1: Both numerator and denominator positive.

So:

$ 7(k - 1) > 0 $ and $ 9 - k > 0 $

First inequality: $ k - 1 > 0 \implies k > 1 $

Second inequality: $ 9 - k > 0 \implies k < 9 $

Therefore, k is integer between 2 and 8 inclusive.

Case 2: Both numerator and denominator negative.

So:

$ 7(k - 1) < 0 $ and $ 9 - k < 0 $

First inequality: $ k - 1 < 0 \implies k < 1 $

Second inequality: $ 9 - k < 0 \implies k > 9 $

But k cannot be both less than 1 and greater than 9, so no solution here. Therefore, only Case 1 is possible: k from 2 to 8 inclusive.

So let me check each integer k from 2 to 8 and see if b is integer and greater than 9.

Let me tabulate:

For k=2:

b = 7*(2 - 1)/(9 - 2) = 7*1/7 = 1. But 1 is not >9. Disregard.

k=3:

7*(3 - 1)/(9 - 3) = 7*2 / 6 = 14/6 = 7/3 ~= 2.333... Not integer. Disregard.

k=4:

7*(4 - 1)/(9 - 4) = 7*3 / 5 = 21/5 = 4.2. Not integer. Disregard.

k=5:

7*(5 - 1)/(9 - 5) = 7*4 / 4 = 28 / 4 = 7. b=7, which is not >9. Disregard.

k=6:

7*(6 - 1)/(9 - 6) = 7*5 / 3 ~= 35/3 ~= 11.666... Not integer. Disregard.

k=7:

7*(7 - 1)/(9 - 7) = 7*6 / 2 = 42 / 2 = 21. b=21, which is >9. So that's valid. So this is one solution.

k=8:

7*(8 - 1)/(9 - 8) = 7*7 / 1 = 49 / 1 = 49. b=49, which is >9. Valid. Another solution.

Therefore, only k=7 and k=8 give valid bases b=21 and 49. Which matches our previous results. Therefore, sum is 21 + 49 = 70. Therefore, same answer.

Therefore, I think that is solid. Therefore, the answer is 70.

But just to make sure, let me check for k=6 again. 35/3 is not integer. Correct. For k=5, 7. Not greater than 9. So yes, only k=7 and 8. Therefore, answer is 70. Therefore, I can be confident.

**Final Answer**
\boxed{70}
\end{lstlisting}

\medskip\noindent\textbf{With \method{} (0.6B), $\alpha{=}5$.}\par\nopagebreak
\begin{lstlisting}
Okay, let me try to figure out this problem. So, we need to find the sum of all integer bases $ b > 9 $ for which $ 17_b $ is a divisor of $ 97_b $. Hmm, let me break this down step by step.

First, I remember that numbers written in different bases convert to base 10 by multiplying each digit by the base raised to the power of its position. For example, $ 17_b $ would be $ 1 \times b + 7 $ in base 10, right? Similarly, $ 97_b $ would be $ 9 \times b + 7 $. Let me confirm that. Yeah, since the rightmost digit is the units place (base^0), and the next is base^1, so yes, that makes sense.

So, translating both numbers to base 10:

- $ 17_b = 1 \times b + 7 = b + 7 $
- $ 97_b = 9 \times b + 7 = 9b + 7 $

The problem states that $ 17_b $ divides $ 97_b $. In other words, $ b + 7 $ divides $ 9b + 7 $. So, mathematically, this means that $ (9b + 7) \div (b + 7) $ should result in an integer. Alternatively, there exists some integer $ k $ such that $ 9b + 7 = k \times (b + 7) $.

Our goal is to find all integer bases $ b > 9 $ for which this divisibility condition holds, then sum those bases.

Let me think about how to approach this equation. Maybe I can express $ 9b + 7 $ divided by $ b + 7 $ as a quotient plus some remainder. If the remainder is zero, then it's divisible. Alternatively, perform polynomial division or manipulate the equation algebraically.

Let me try to perform the division. Let me write $ \frac{9b + 7}{b + 7} $. Let's do polynomial division or simplify it.

First, divide 9b by b, which is 9. Multiply (b + 7) by 9: that gives 9b + 63. Subtract that from 9b + 7:

(9b + 7) - (9b + 63) = 9b + 7 - 9b - 63 = -56.

Therefore, $ \frac{9b + 7}{b + 7} = 9 - \frac{56}{b + 7} $. 

For this to be an integer, $ \frac{56}{b + 7} $ must be an integer. Therefore, $ b + 7 $ must be a positive divisor of 56.

But since $ b > 9 $, then $ b + 7 > 16 $. Therefore, we need to find all divisors of 56 that are greater than 16, then subtract 7 to get the corresponding bases, and sum them up.

Wait, let me check. If $ b + 7 $ divides 56, then $ b + 7 $ is a positive divisor of 56. However, since $ b > 9 $, then $ b + 7 > 16 $. Therefore, we need to find all divisors of 56 that are greater than 16, then subtract 7 from each of them to get valid bases $ b $, then sum those bases.

First, let me list all positive divisors of 56. The prime factorization of 56 is $ 2^3 \times 7^1 $. Therefore, the divisors are:

1, 2, 4, 7, 8, 14, 28, 56.

So divisors greater than 16 are 28 and 56. Therefore, $ b + 7 = 28 $ or $ b + 7 = 56 $. Therefore:

If $ b + 7 = 28 $, then $ b = 21 $.

If $ b + 7 = 56 $, then $ b = 49 $.

Therefore, the bases are 21 and 49. Their sum is 21 + 49 = 70.

Wait, is that all? Let me verify if there are any other divisors. The divisors of 56 are indeed 1, 2, 4, 7, 8, 14, 28, 56. The next one after 14 is 28, which is greater than 16. So yes, only 28 and 56. Therefore, bases 21 and 49, sum is 70.

But let me check with actual numbers to make sure.

For base 21:

$ 17_{21} = 1*21 + 7 = 28 $.

$ 97_{21} = 9*21 + 7 = 189 + 7 = 196 $.

Check if 28 divides 196: 196 / 28 = 7. Yes, that's an integer.

For base 49:

$ 17_{49} = 1*49 + 7 = 56 $.

$ 97_{49} = 9*49 + 7 = 441 + 7 = 448 $.

Check if 56 divides 448: 448 / 56 = 8. Yes, that's also an integer. Perfect.

Therefore, the bases are 21 and 49, sum is 70. So the answer is 70.

But let me just make sure there isn't a negative divisor or something? But since $ b > 9 $, and bases are positive integers greater than the digits they contain. Since the digits here are 9 and 7, the base has to be greater than 9. So base must be at least 10, but we have the condition b > 9, so starting from 10 upwards. But since we have b + 7 divides 56, and divisors of 56 are positive, as negative divisors would lead to negative b + 7, which would make b negative, which isn't allowed. So only positive divisors. Therefore, yes, 28 and 56 are the only ones. Hence, 21 and 49. Sum is 70.

Therefore, the answer is \boxed{70}.

**Final Answer**
\boxed{70}
\end{lstlisting}

\end{document}